\documentclass[journal,compsoc]{IEEEtran}
\usepackage{times,enumerate,caption,algorithmic}
\usepackage{multirow,booktabs,supertabular,balance,array,makecell,amssymb,balance,epsfig}
\usepackage[breaklinks=true,colorlinks,bookmarks=false]{hyperref}
\usepackage{dblfloatfix}
\usepackage[latin1]{inputenc}
\usepackage[english]{babel}
\usepackage[table,xcdraw]{xcolor}
\usepackage{rotating}
\usepackage{tabularx}
\usepackage{amsfonts}
\usepackage{soul}
\usepackage{csquotes}
\usepackage{setspace,amsthm,bm,amsmath}
\usepackage{pgfarrows}
\usepackage[flushleft]{threeparttable}
\usepackage{xspace}
\usepackage{enumitem}
\usepackage[edges]{forest}
\usetikzlibrary{shadows.blur}

\newcommand{\latinphrase}[1]{\textit{#1}} 
\newcommand{\etal}{\latinphrase{et~al.}\xspace}
\newcommand{\ie}{\latinphrase{i.e.}\xspace}

\newcommand{\etc}{\latinphrase{etc.}\xspace}

\begin{document}

\title{A Survey on Image Aesthetic Assessment}
\author{Abbas Anwar$^*$, 
        Saira Kanwal$^*$, 
        Muhammad Tahir, 
        Muhammad Saqib, \\
        Muhammad Uzair, 
        Mohammad~Khalid~Imam~Rahmani, 
        Habib Ullah
\IEEEcompsocitemizethanks{
\IEEEcompsocthanksitem Abbas Anwar is research assistant with University of Engineering and Technology, Peshawar.\protect\\ 

\IEEEcompsocthanksitem Saira Kanwal is research assistant with COMSATS University Islamabad, Wah Campus, Pakistan\protect\\ 

\IEEEcompsocthanksitem Muhammad Tahir  is associate professor are  with Saudi Electronic University, Saudi Arabia.\protect\\
Corresponding Author: Muhammad Tahir (m.tahir@seu.edu.sa)\protect\\ 

\IEEEcompsocthanksitem Muhammad Saqib is research scientist with Imaging and Computer vision group, Data61-CSIRO, Australia.\protect\\

\IEEEcompsocthanksitem Muhammad Uzair is assistant professor with COMSATS University Islamabad, Wah Campus, Pakistan\protect\\ 

\IEEEcompsocthanksitem Mohammad~Khalid~Imam~Rahmani are assistant professor with Saudi Electronic University, Saudi Arabia.\protect\\

\IEEEcompsocthanksitem Habib Ullah is associate professor scientist Norwegian University of Life Sciences, Norway.\protect\\
\IEEEcompsocthanksitem $^*$ denotes equal contribution.\protect\\}
}
\markboth{Journal}%
{Anwar \MakeLowercase{~\etal}: A Survey on Image Aesthetic Assessment}
\maketitle

Automatic image aesthetics assessment is a computer vision problem dealing with categorizing images into different aesthetic levels. The categorization is usually done by analyzing an input image and computing some measure of the degree to which the image adheres to the fundamental principles of photography such as balance, rhythm, harmony, contrast, unity, look, feel, tone and texture. Due to its diverse applications in many areas, automatic image aesthetic assessment has gained significant research attention in recent years. This article presents a review of the contemporary automatic image aesthetics assessment techniques. Many traditional hand-crafted and deep learning-based approaches are reviewed, and critical problem aspects are discussed, including why some features or models perform better than others and the limitations. A comparison of the quantitative results of different methods is also provided.
\section{Introduction}
\begin{figure}[t]
\centering
\begin{tabular}{cc} 
    \includegraphics[width=0.45\columnwidth]{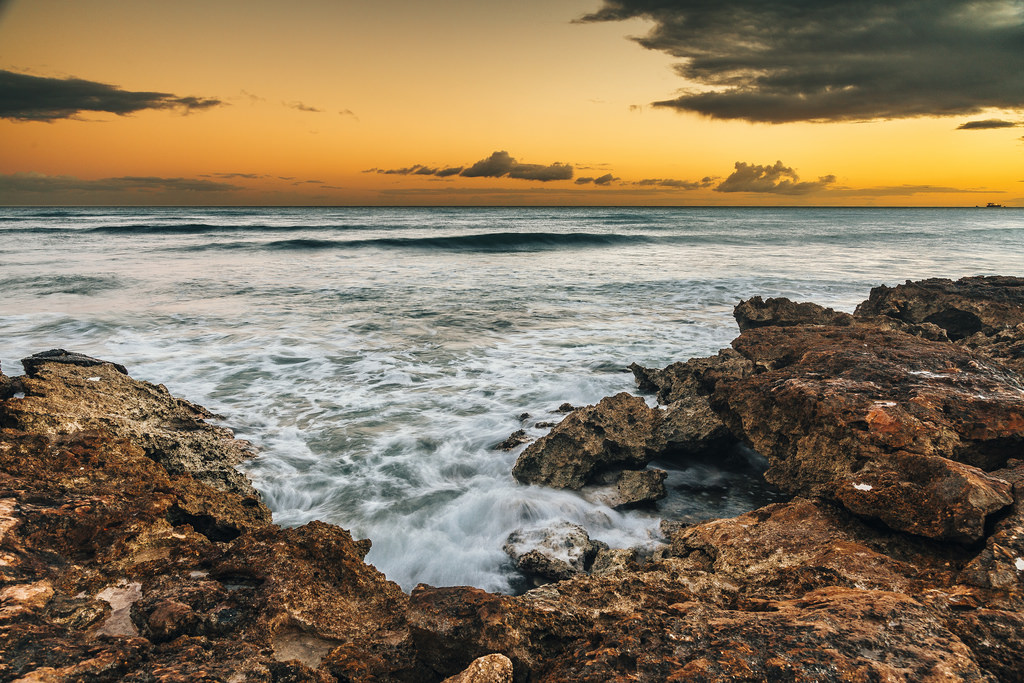}&
   	\includegraphics[width=0.45\columnwidth]{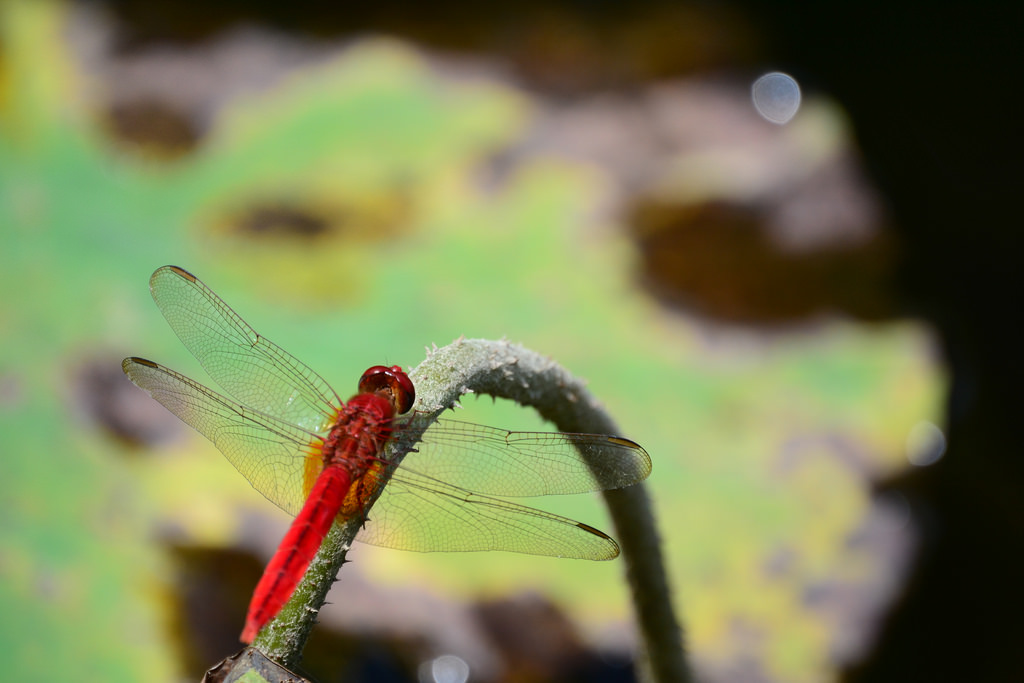}\\
	a) &	b) \\
	 
	\includegraphics[width=0.45\columnwidth]{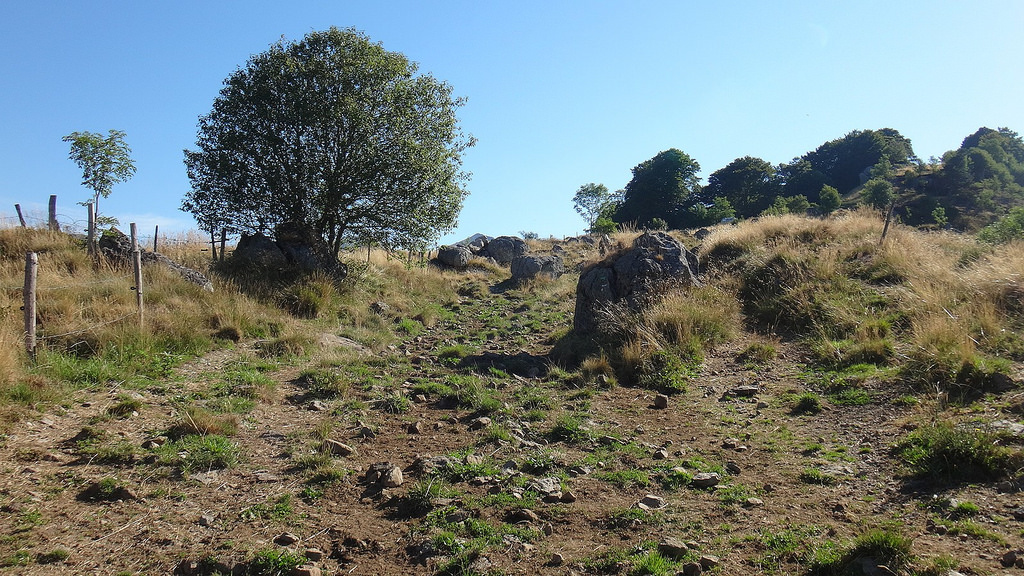}&
	\includegraphics[width=0.45\columnwidth]{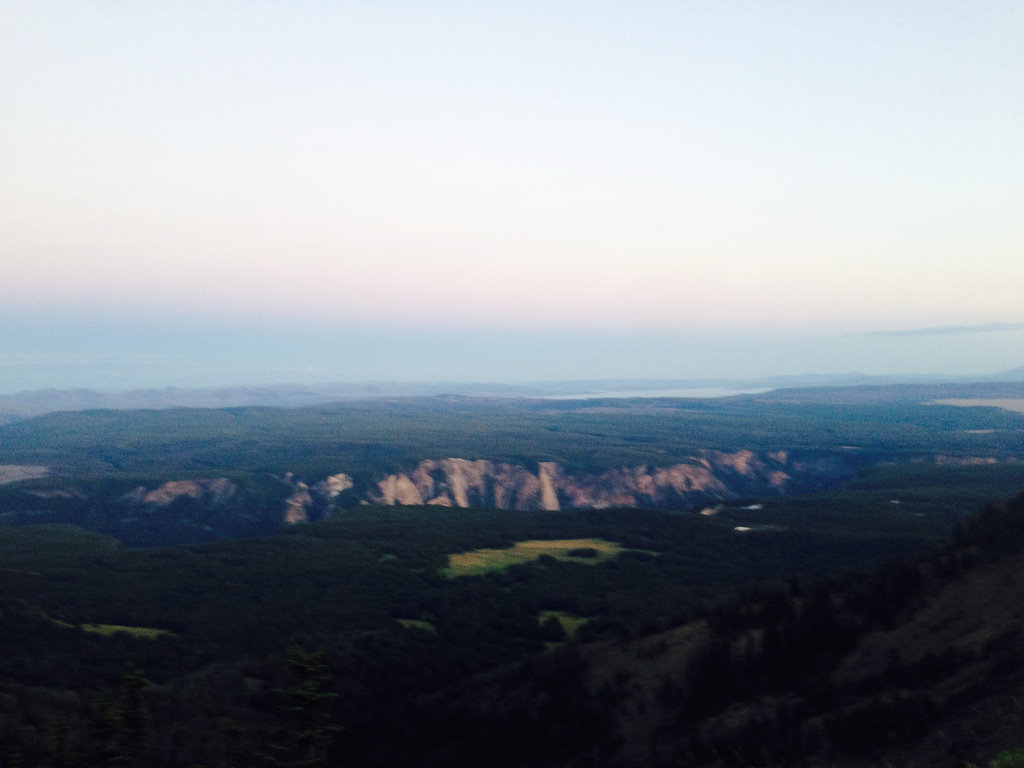}\\
	c) & 	d)\\

\end{tabular}
\caption{{The sample images are taken from the  Aesthetics and Attributes Database (AADB)~\cite{kong2016photo}, consisting of various photographic imagery of real scenes collected from Flickr. For each image, the rating is provided by averaging five rater's score as a ground-truth score. Eleven aesthetic attributes were considered while curating the dataset, such as \emph{interesting content, object emphasis, good lighting, colour harmony, vivid colour, shallow depth of field, motion blur, rule of thirds, balancing element, repetition, and symmetry.} The photos a) and b) represent photos with a high aesthetic score of 1.0 and 0.7, while c) and d) represent the photos with a low aesthetic score of 0.4 and 0.1, respectively. }}
\label{fig:aesthetics_show}
\end{figure}

It may be true that beauty lies in the eyes of the beholder but for a computer, automatically quantifying the beauty of a photograph is a challenging task. 
In computer vision, the task is known as automatic image aesthetics assessment and deals with quantifying the beauty, quality, and impression of photographs to categorize images into different aesthetic levels as shown in Figure~\ref{fig:aesthetics_show}. Image aesthetic assessment has diverse applications in multimedia content generation, processing, and communication. For example, it can be employed to benchmark the algorithms for image noise removal and image restoration as well as for monitoring of quality of service (QoS) in systems where images are digitally compressed, communicated, and decompressed.

\begin{figure*}[t]
\centering
\begin{forest}
forked edges,
for tree={draw,align=center,edge={-latex},fill=white,blur shadow}
[Hand-Crafted Methods
 [Basic Methods
    [Lo~\etal~\cite{ref1}\\
    Datta~\etal~\cite{ref3}\\
    Redi~\etal~\cite{ref9}\\
    Aydin~\etal~\cite{ref11}\\
    Mavridaki~\etal~\cite{ref8}\\
    Li~\etal~\cite{ref2}\\
    Domen~\etal~\cite{ref13}]
 ]
 [Statistical Methods
    [Yang~\etal~\cite{ref4}\\ 
    Bhattacharya~\etal~\cite{ref14}\\
    Bhattacharya~\etal~\cite{ref34}\\
    Lo~\etal~\cite{ref35}\\
    Wang~\etal~\cite{ref41}]
 ]
 [Local $\&$ Global Features
    [Goa~\etal~\cite{ref7}\\
    BLIINDS-II~\cite{ref43}\\
    SDAM~\cite{ref10}\\
    Wang $\&$ Simoncelli~\cite{ref44}\\
    Riaz~\etal~\cite{ref45} ]
 ]
 [Content Based Approaches
    [Nishiyama~\etal~\cite{ref6}\\
    Marchesotti~\etal~\cite{ref12}\\
    CPQA~\cite{ref36}\\
    PGIA~\cite{ref37}\\
    Su~\etal~\cite{ref38}\\
    Rongju~\etal~\cite{ref39} ]
 ]
]
\end{forest}
\vspace{1mm}
	\caption{Overview and Categorization of hand-crafted based techniques for image aesthetics assessment.}
	\label{fig53}
\end{figure*}
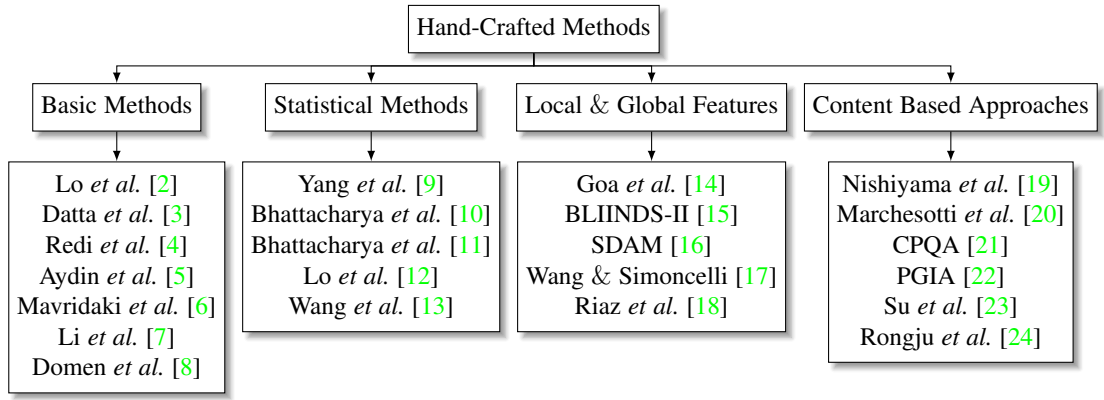

Image recognition and classification systems can also benefit from aesthetic assessment techniques~\cite{ref30}. For instance, in biometric systems~\cite{uzair2017non}, image quality assessment can be used, and image enhancement approaches can be applied to improve quality and accuracy in case of low-quality image as input~\cite{ref31}\cite{ref32}~\cite{ref33}. Moreover,  image quality assessment can be utilized in robotics, where a robot automatically assesses the image quality and change focus and position to recapture the image if the quality metric is below some recommended level. Due to its significant application potential in the rapidly increasing digital camera and photography industry, automated image aesthetic assessment has recently gained considerable research attention from the computer vision and pattern recognition community~\cite{ref25,ref26,ref27,ref28,ref29}. 
Automatic image aesthetic assessment has many challenges. For example, the input visual data may contain noise and image artifacts such as illumination and environmental conditions. Focus and pose deflections introduce disparities in images. Images may be subject to variations in colour harmony because of sensor resolution issues. Background clutter can also hinder the accuracy of aesthetic assessment algorithms. Moreover, the visual judgment conflicts of humans also translate to different challenges for image quality rating algorithms.

Over the past couple of decades, many computer vision techniques have been developed for image aesthetic/quality assessment. Both hand-crafted feature-based approaches and deep learning-based approaches have been exploited for the task. Hand-crafted features-based algorithms generally design filters to encode aspects of the image aesthetics such as photographic rules, image texture, local and global content features, \etc The represented aesthetics features are then fed to classical machine learning approaches to classify the image in different aesthetic levels. Deep learning-based techniques use robust deep neural networks to learn and encode image aesthetics from a large number of training images. Deep learning-based methods are more accurate as they can model more complex image features and their relationships. 

This article provides a survey of techniques for automatic image aesthetic assessment. Both hand-crafted feature-based methods and the recent deep learning approaches are covered in detail, describing each technique's basic framework with its pros and cons. 
The outcomes of experimental method in terms of accuracy, the dataset used, its size, and the depth of each aesthetic rating algorithm are also discussed.

\vspace{1mm}
\noindent
\textbf{Motivation}: We would like to emphasize that a survey is required in image aesthetics due to many papers published in deep learning; although a review of image aesthetics~\cite{deng2017image} was published half a decade ago, we argue that the number of articles published in the last year is significantly higher during the previous five years, therefore we expect further increase in the coming years. Furthermore, the review of~\cite{deng2017image} is more on the lines of explaining the image aesthetics and lacks in listing hand-crafted or deep learning methods in detail. Similarly, we want to provide detailed descriptions of important articles to help the community adopt the most appropriate approach and avoid reproducing the methodologies. Likewise, we strive to give a good research direction through this survey, specify the gaps and limitations, and provide future direction.

\section{Hand-Crafted Methods}

Although hand-crafted features are considered a thing of the past, they still provide good insight into a computer vision task. Hand-crafted methods primarily design some kinds of pixel filters to extract or encode low-level image features. Standard features used by the hand-crafted techniques include colour, contrast, saturation, brightness, texture and foreground-background statistics, global features, and local features ratio statistics~\cite{mahmood2018multi}. Figure~\ref{fig53} summarizes different techniques based on the features they use to encode the aesthetic information about images. We discuss each of these categories in detail in the following sections.

\subsection{Basic Methods}
These methods are pioneering image aesthetic approaches and provide a naive methodology for accuracy. 

\begin{enumerate} [align=left,   leftmargin=0em,   itemindent=5pt,   labelsep=6pt,   labelwidth=0em ]
\setlength\itemsep{0.3em}

\item An intelligent photographic interface is proposed by Lo~\etal~\cite{ref1}, with on-device aesthetic quality assessment for bi-level image quality on general portable devices~(Figure~\ref{fig:basic_methods}(a)). In this framework, photographic rules were followed, and a three-layered structure was designed. Using hand-tuned techniques, the first layer extracted composition, saturation, colour combinations, contrast, and richness features. In the second layer, an independent SVM classifier~\cite{ullah2019internal} was trained for each feature perspective to obtain the feature index. Moreover, the SVM classifier is trained to get the aesthetic score in the last layer. The mentioned framework is tested on CUHK~\cite{yan2013learning} dataset, comprising 2078 high-quality and 7573 low-quality images, providing an accuracy of 89\%.

\item A computational algorithm using region-based features and k-means clustering is presented by Datta~\etal~\cite{ref3}. Colour segments are extracted from the image utilizing region-based features and texture information to assess the quality of images with the connected component technique. Subsequently, the SVM classifier on the extracted feature is trained to categorize images into high and low aesthetic categories. A regression model~\cite{uzair2015hyperspectral} is also trained to obtain a regression score. The dataset is collected from a photo-sharing website consisting of 3581 images.   
 
\begin{figure*}[t]
\centering
\begin{tabular}{cc} 
    \includegraphics[width=\columnwidth]{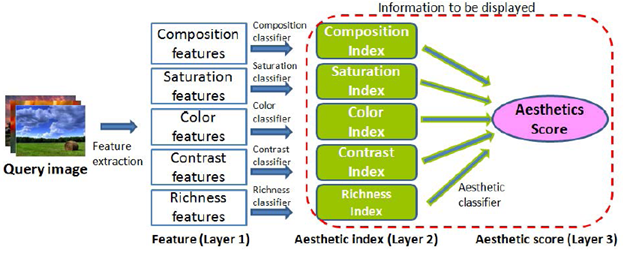}&
   	\includegraphics[width=0.6\columnwidth]{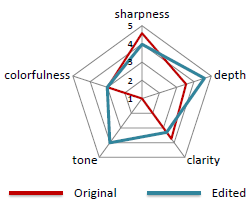}\\
	a) & b)\\
	\includegraphics[width=0.8\columnwidth]{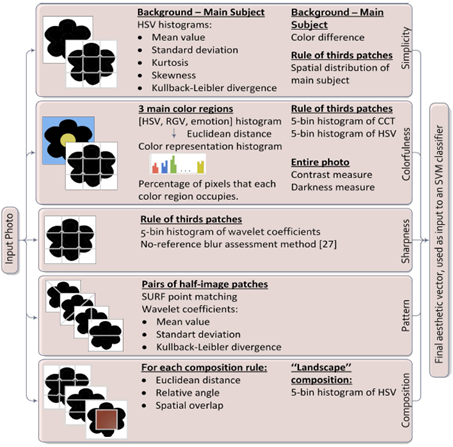}&
	 \includegraphics[width=\columnwidth]{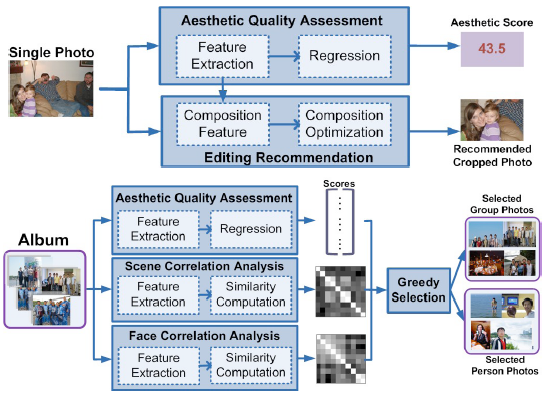}\\
	c) & d) \\	
	
\end{tabular}
\caption{Comparative analysis of Basic hand-crafted assessment methods. a) Hierarchical Aesthetic Assessment Algorithm~\cite{ref1},
	b) Proposed Attributes in~\cite{ref11},	c) Rule Based System~\cite{ref8},
	d) Photo Quality Assessment $\&$ Selection System~\cite{ref2}. }
\label{fig:basic_methods}
\end{figure*}
 
\item To access the quality of digital portraits, Redi~\etal~\cite{ref9} introduced a technique based on composition, scene semantics, portrait-specific features, correct perception of signal and fuzzy properties, and the five essential features extracted from images. One should note that composition rules are the essential and basic photography rules, including sharpness, spatial arrangement, lighting, texture, and colour. The semantic contents represent the overall photography depiction, including high-level features~\cite{ullah2019stacked}. The correct perception of signals includes noise, contrast quality, exposure quality, and JPEG quality, while portrait-specific features include face position, face orientation, age, gender, eye, nose, mouth position, foreground, and background contrast. Fuzzy properties are originality, memorability, uniqueness, and emotion depiction. LASSO regression~\cite{ref48} is applied to the extracted composition features, learning regression parameters for every feature group. Moreover, a correlation between the predicted score and the original aesthetic value is computed. Using regression on all features, a final aesthetic score is predicted. First, the framework is tested on a small scale, and later it is tested on a large scale, classifying the images as beautiful and non-beautiful via SVM classification. The AVA dataset is used for training and testing, achieving 75.76\% accuracy.

\begin{figure*}[t]
\centering
\begin{tabular}{cc} 
    \includegraphics[width=\columnwidth]{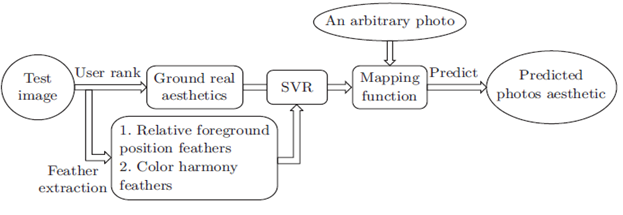}&
	\includegraphics[width=0.8\columnwidth]{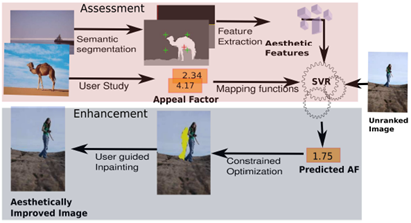}\\
	a) The Framework of Yang~\etal~\cite{ref4}&
	b) The Method of Bhattacharya~\etal~\cite{ref14}\\
\end{tabular}
\caption{Comparative analysis of hand-crafted statistical methods.}
\label{fig:statistical_methods}
\end{figure*}

\item A photographic rating framework that computes aesthetic signatures using attributes of colorfulness, sharpness, depth, tone, and clarity as shown in Figure~\ref{fig:basic_methods}(b) is introduced by Aydin~\etal~\cite{ref11}. Generally, relation to photographic rules and clear definition such as sharpness, clarity, colourfulness, \etc are essential building blocks of any image's aesthetic algorithm. In the framework, a picture is shown on a screen with five images displayed on other screens and a short task description to determine the stimuli from the image for the five primary attributes, \ie colourfulness, sharpness, depth, tone, and clarity where the task description contains an aesthetic rating of the image individually for each attribute, working on 8-bit RGB images. The algorithm works in three steps: i) convert the input image to a double-precision image and normalize it, ii) an edge pyramid is computed with domain transform applied to each pyramid layer, and iii) a multi-scale~\cite {khan2019disam} contrast image is estimated. Moreover, a data structure is built using detailed contrast images, known as a focus map that indicates in-focus regions in the image, and the inverse of the focus map depicts out-of-focus image regions. The focus map is used to calculate features such as depth, colourfulness, sharpness, clarity, and tone. The training is performed on 955 images randomly selected from DpChallange~\cite{27} dataset. This research is applied mainly in HDR tone mapping, automatic photo editing applications, auto aesthetic analysis, and multi-scale contrast manipulation.

\item Mavridaki~\etal~\cite{ref8} introduced a system using five basic photography rules: simplicity, colourfulness, sharpness, pattern, and composition. The \emph{simplicity} refers to capturing images with emphasized subjects. For \emph{colourfulness}, k-means clustering is performed to separate different colours. For \emph{sharpness}, blur detection algorithm~\cite{89} is employed, and for \emph{pattern} assessment, SURF point features~\cite{ref47} are extracted.  For \emph{composition} rule, landscape compostion~\cite{90}, and rule of thirds~\cite{ref74}  are examined. All these features are combined in the last stage to produce an element feature vector fed to an SVM classifier and are depicted in Figure~\ref{fig:basic_methods}(c). The mentioned method is evaluated on 12k images collected from CUHKPQ~\cite{luo2011content}, CUHK~\cite{yan2013learning}, and AVA~\cite{murray2012ava} datasets, where half of them are high-quality, and the other half are of low-quality images. The proposed framework achieves an overall accuracy of 77.08 \%.

\item An online photo-quality assessment and photo selection system is present in~\cite{ref2} as shown in Figure~\ref{fig:basic_methods}(d), where the users post their images, and the algorithm provides aesthetic evaluation and editing recommendations. The cropping-based editing algorithm uses composition features and composition optimization for the proposed system inputting a single image or photo album. The aesthetic score is calculated for a single image between 0-100, and image crop recommendations are provided if the aesthetic score is less than 70. Similarly, the top ten rated group photos and single-person photos are displayed in the photo album with respective scores. 
Aesthetic assessment is performed by extracting a feature vector from images followed by regression to compute the aesthetic score. Features are considered based on colour, light, composition, and face characteristics. The aesthetic quality assessment algorithm is trained on a dataset of 500 photos collected from Amazon Mechanical Turk with ten test images. For albums, images are categorized into single-person and group photos. Moreover, scene correlation analysis and face correlation analysis are performed for group photos, and aesthetic quality is determined.

\item Using feature extraction and SVM classifier, Domen~\etal~\cite{ref13} proposes an aesthetic photo technique, where three basic photography features, including simplicity, composition, and colour selection, are considered for aesthetic assessment. The edge features determine simplicity and the ratio of background to image colour palette. The rule-of-thirds and golden ratio assess composition. To classify image in the high aesthetic score and low aesthetic score, the SVM classifier is trained on 258, and 1048 images are randomly selected from the Flicker and the DPChallange~\cite{27} datasets, respectively, achieving an accuracy of 95\% using 73 features from each image.

\end{enumerate}

\subsection{Statistical Methods}
In this subsection, we discuss various methods for image aesthetics based on the statistics of texture, foreground and background.

\begin{enumerate}[align=left,   leftmargin=0em,   itemindent=5pt,   labelsep=6pt,   labelwidth=0em]
\setlength\itemsep{0.3em}

\item The landscape photo assessment algorithm by Yang~\etal~\cite{ref4} is shown in Figure~\ref{fig:statistical_methods}(a). The authors extract as relative foreground position and colour harmony features, and according to the rule of thirds, the object of interest must be at the image centre. Moreover, colour harmony is the relative position of each colour in the spatial domain, and colour harmonic normalization \cite{ref49} is performed via hue wheel. The support vector regression (SVR) algorithm~\cite{ref50} is trained to map the foreground position and colour harmony features with the ground real aesthetics. A mapping model is learned to predict the aesthetic level after achieving the composition deviation and is tested on 431 images from Pconline and Flickr~\cite{91}  concerning 84.83\% accuracy.



\item Recently, a photo-quality assessment and enhancement algorithm to train SVR employing the relative foreground and visual weight ratio image features is given in~\cite{ref14}, the architecture of their proposed framework is in Figure~\ref{fig:statistical_methods}(b). The image is edited if the appeal factor is lower than the computed aesthetic score (\ie between 1 to 5). The dataset consists of 384 single object images, and 248 images are scene images downloaded from  Flickr~\cite{91}. This approach achieves 86\% accuracy.

\item After image and photo aesthetic assessment~\cite{ref14}, Bhattacharya~\etal~\cite{ref34} next presented an aesthetic assessment framework for videos. As the algorithm deals with videos, three-level features are extracted, including cell level, frame level, and shot level. The \emph{cell features} comprise dark-channel, sharpness, and eye sensitivity. For \emph{frame-level}, Sentibank library~\cite{121} detects 1,200-dimensional feature vector. For \emph{shot features}, the foreground motion~\cite{ullah2019single,ullah2020multi,ullah2018anomalous,ullah2017density}, the background motion and the texture dynamics~\cite{ullah2019hybrid} are computed from the video. A SVM is trained for each mentioned level feature. Finally, all SVM scores are fused using low-rank late fusion (LRLF)~\cite{ref51} while the algorithm is evaluated using NHK dataset~\cite{122} comprising of 1k videos.  

\item Lo~\etal~\cite{ref35} utilizes the colour palette, layout composition, edge composition, and global texture features for aesthetic assessment. The HSV histogram colour components extract colour palette features; layout composition features are determined through the $\ell_1$ distance between H, S, and V channels; edge detection filters compute edge composition features~\cite{uzair2020bio}; global texture features are calculated by the sum of absolute differences between four channels. In addition to the features mentioned above, blur, dark channel, contrast, and HSV counts are also computed. An SVM classifier trained on the CUHK dataset rates the image in high and low aesthetic levels, providing 86\% accuracy in the performance.

\item Using saliency enhancement, Wang~\etal~\cite{ref41} introduced an image aesthetic level prediction algorithm. The authors use the salient region of the image to represent objects, computing the saliency map via Itti\'s visual saliency model \cite{ref42}. The visual features from the image are extracted \ie global, saliency regions, and foreground-background relationship features, where the global features are composed of texture details, low depth of fields, and rule of thirds. It is also to be noted that the distribution, position, and area of salient regions are determined as features of salient regions. The hue count and edge spatial distribution represent the foreground-background relationship. Moreover, images are classified into high and low aesthetic levels by an SVM classifier trained on a dataset downloaded from Photo.net~\cite{123}, which contains 3161 images and achieves an accuracy of 83.7\%. 

\end{enumerate}

\begin{figure*}[t]
\centering
\begin{tabular}{c@{  }c} 
    \includegraphics[width=0.45\textwidth]{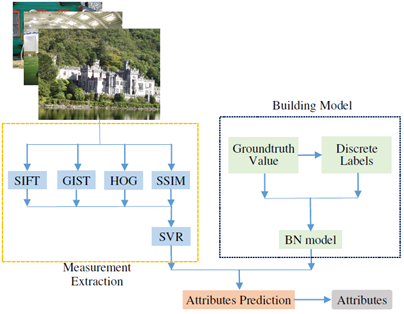}&
    \includegraphics[width=0.45\textwidth]{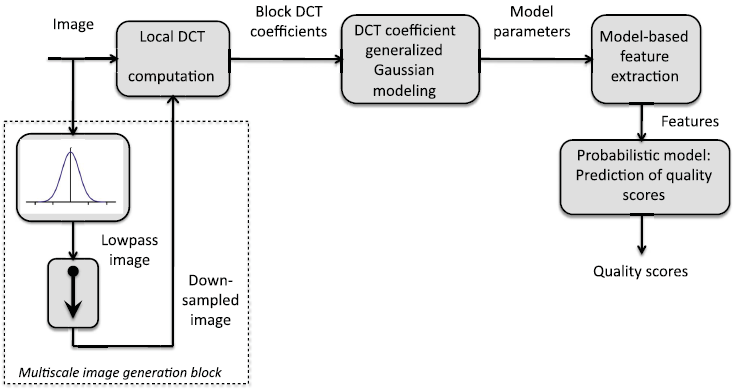}\\
   
    a) Gao~\etal~\cite{ref7}&
    b) The BLIINDS-II algorithm~\cite{ref43}\\ 
    
   	\includegraphics[width=0.45\textwidth]{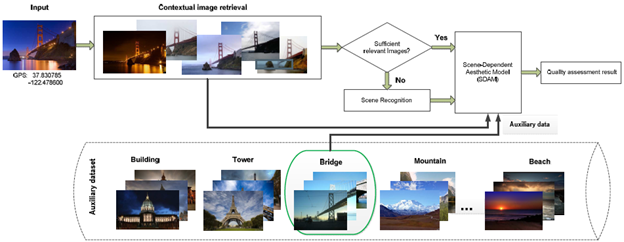}&
   	\includegraphics[width=0.45\textwidth]{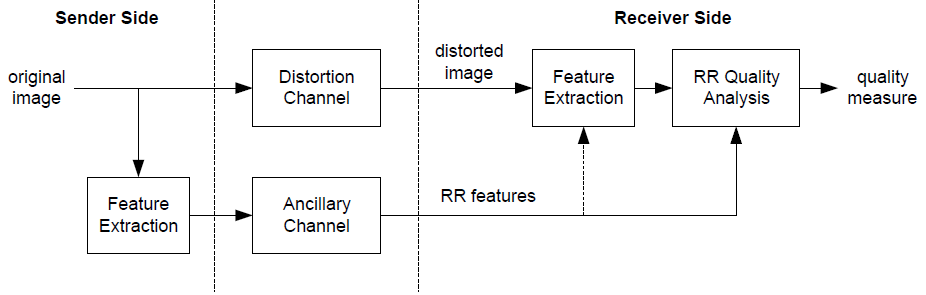}\\
    c) Yin~\etal~\cite{ref10} &
    d) Wang $\&$ Simoncelli~\cite{ref44} \\
\end{tabular}
\caption{The frameworks of local and global hand-crafted features algorithms.}
\label{fig:Local_Global}
\end{figure*}

\subsection{Local and Global Features Methods}
In this section, we summarize the algorithms that consider both local and global features learned from images for aesthetic assessment.

\begin{enumerate}[align=left,   leftmargin=0em,   itemindent=5pt,   labelsep=6pt,   labelwidth=0em ]
\setlength\itemsep{0.3em}

\item The multi-label task for assessing the aesthetic quality of images based on different aesthetic attributes like aesthetic, memorable, and attractive attributes using high-level semantic information is explored in~\cite{ref7} as shown in Figure~\ref{fig:Local_Global}(a) by designing a Bayesian Network to predict the aesthetic level using multi aesthetic attribute prediction. Furthermore, a three-node Bayesian Network presents each aesthetic attribute, including its label, value, and measurement. There are two modules of the mentioned framework measurement acquisition by SIFT~\cite{ref53}, GIST~\cite{ref54}, HOG \cite{ref55} or self-similarities and multi-attribute relation modeling. Finally, a support vector regression (SVR) is trained, the ground truth values are discretized in the building model, and a hybrid Bayesian Network structure is learned on continuous and discrete values. The training (with ten-fold cross-validation) and testing are performed on the memorability dataset~\cite{124} containing 2222 images and are evaluated on three different metrics: F1-score, Kappa, and accuracy.

\item The BLIINDS-II algorithm~\cite{ref43} employs discrete cosine transform (DCT)~\cite{ref56,uzairDCT} is given in Figure~\ref{fig:Local_Global}(b), where local DCT is computed utilizing input image and lowpass downsampled image. Afterwards, a gaussian model is built, extracting model-based features, which are then fed to a Bayesian model that predicts the quality scores. The simple Bayesian probabilistic model requires minimum training~\cite{125} and is trained on randomly selected data samples from the LIVE IQA dataset~\cite{126} containing 779 images. The algorithm yields 91\% accuracy.

\item A scene-dependent aesthetic model (SDAM)~\cite{ref10} takes into account both visual content and geo-context by utilizing the transfer learning~\cite{uzair2016blind} approach, where input images along with their geo-context (online images with similar contents as that of the input image) are used (see Figure~\ref{fig:Local_Global}(c)). The SDAM learns from two types of images, \ie one category is geo contextual images that are location-wise similar to online photos, and in the other category, similar class images from the available database (DB). If a sufficient number of contextual images are available, the machine learning~\cite{uzair2018representation} approaches are applied to access the input image quality. The contextual image retrieval may contain the location of the same images but with different objects where the GIST identifies these types of irrelevant images and are discarded. Moreover, to learn, SDAM uses a state vector machine (SVM), which is tested on 9600 geo-tagged and 32k auxiliary dataset images, achieving an accuracy of 81\% on popular spots and 73\% accuracy on images of less prominent locations.

\item Wang $\&$ Simoncelli~\cite{ref44} uses a wavelet domain natural image statistical model, providing a distortion measure algorithm for communication systems where images are transferred from one location to other. The input image is decomposed into 12 wavelet bands, \ie three scales and four orientations. The six wavelet bands are randomly selected to extract features and minimize KLD~\cite{ref57}, rendering a quality score to rate images in different distortion levels. The architecture of the proposed deployment scheme is given in Figure~\ref{fig:Local_Global}(d). The framework is tested on a LIVE database containing 489 images showing 92\% accuracy.

\item Recently, Riaz~\etal~\cite{ref45} employs generic features, including both global and local features, by extracting SURF features in addition to wavelet and composition features. The method also determines basic photographic features, colour combination, saturation, contrast, smoothness, intensity, hue, and aspect ratios from the input image. The approach applied three steps; in the first step, the online database comprising 250 images ( downloaded from Photo.net), in the second step, human professionals rate pictures, and the third step, all the features mentioned above are extracted. An artificial neural network is trained on these features, achieving 83\% accuracy.

\end{enumerate}
 

\begin{figure*}[t]
\centering
\begin{tabular}{cc} 
   \raisebox{5pt}{\includegraphics[width=\columnwidth]{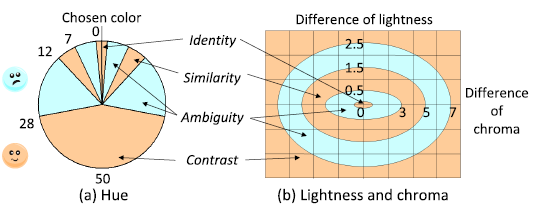}}&
   	\includegraphics[width=0.8\columnwidth]{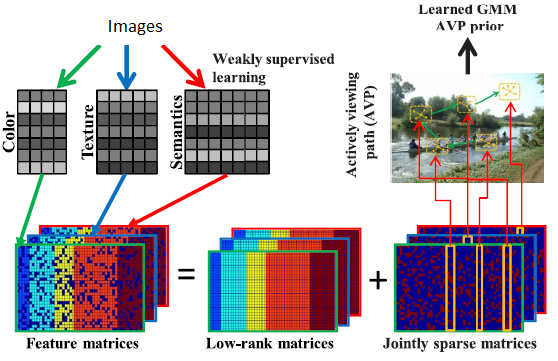}\\
	a)&b)\\
	
\multicolumn{2}{c}{\includegraphics[width=\columnwidth]{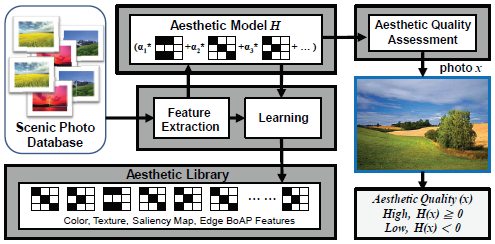}}\\
\multicolumn{2}{c}{ c)}\\

\end{tabular}
\caption{Comparative analysis of Content-Based Methods. 	a) Nishiyama~\etal~\cite{ref6}'s Moon and Spencer Model, 
	b) Algorithm by Zhang~\etal~\cite{ref37}, and c) Su~\etal~\cite{ref38}'s 	algorithm overview}
\label{fig:content_methods}
\end{figure*}

\subsection{Content-Based Methods}
Content-based methods take into account the content of the images. We provide an overview of such methods in the following paragraphs.

\begin{enumerate}[align=left,   leftmargin=0em,   itemindent=5pt,   labelsep=6pt,   labelwidth=0em ]
\setlength\itemsep{0.3em}

\item Aesthetic quality is highly based on the local region's sum of colour harmony scores according to Nishiyama~\etal~\cite{ref6}, implementing bag-of-colour patterns for photograph quality classification (see Figure~\ref{fig:content_methods}(a)). The authors employ the moon and spencer model~\cite{ref62,ref63} computes the sum of colour harmony scores. The colour model evaluates the hue, chroma, and lightness from the sampled local regions of images, and then the collected distributions are integrated to form a bag-of-features \cite{ref61} framework. Every local area is described using simple colour patterns of colour harmony models, assuming the colour distribution to be simple. Aesthetic rating is classified by calculating the histogram of each colour pattern. The SVM classifier is trained on 124,664 images collected from the DPChallange~\cite{27} dataset to predict the photograph quality, categorized into high and low aesthetic levels. The algorithm is tested in two scenarios: a) whole image and 2304 local regions each of size 32 $\times$ 32, and (b) absolute and relative colour values offer an overall  77.6\% accuracy. To further improve the classification,  the authors also consider the saliency, blur, and edge features in addition to colour harmony patterns.

\item Marchesotti~\etal~\cite{ref12} proposed an image descriptor-based with Fisher vector (FV)~\cite{ref64} and bag-of-visual-words (BOV)~\cite{ref65} which extracts generic descriptors from image and gradient information is obtained through Scale Invariant Feature Transform (SIFT). The input image is divided into patches, and for each patch, BOV computes discrete distribution, and FV calculates continuous distribution. Furthermore, SIFT is applied to each patch, and GIST descriptor is also considered, initially designed for scene categorization. The algorithm is evaluated on two datasets, Photo.net and CHUK, consisting of 3581 images and  12k images, respectively. The BOV and FV features are computed from 32$\times$32 patches at five different scales and represented by SIFT that generates a 128-dimensional feature vector for each patch reduced to 64 dimensions using PCA. The EM~\cite{mclachlan2007algorithm} algorithm learns visual vocabulary Gaussian mixture models, and the SVM classifier is learned using hinge loss and stochastic gradient descent algorithm \cite{ref68, ref69}. In their experiments, Fisher Vector outperforms all other techniques and delivers a maximum of 78\% accuracy.

\item The content-based photo quality assessment, abbreviated as (CPQA)~\cite{ref36}, deals with both regional and global features concerning three different areas, including clarity-based detection, layout-based detection, and human-based detection. Regional features extracted from the input image are dark channel, clarity-contrast, lighting-contrast, composition geometry, complexity, and brightness. Besides, the global features include hue and scene composition features. An SVM is trained on the CUHK-PQ~\cite{luo2011content}, including 17673 images classifying the images into high, low, and uncertain categories. The CPQA algorithm gives 83\% accuracy.

\item The perception-guided image aesthetic (PGIA)~\cite{ref37} assessment algorithm learns the model constrained with different low-rank graphlets created by fusing low-level and high-level features from the image. The sparsity of the graphlets is then calculated to generate jointly sparse matrices as shown in Figure~\ref{fig:content_methods}(b). The mentioned graphlets turn into actively viewing path (AVP) descriptors, and the Gaussian Mixture Model learns the distribution of these aesthetic descriptors. The proposed algorithm is trained and tested on AVA~\cite{murray2012ava}, Photo.net, and CUHK datasets comprising of 12k, 3581 images, and 25k images, providing 90.59\%, 85.52\%, and 84.13\% accuracy, respectively.

\item Su~\etal~\cite{ref38} proposes a bag-of-aesthetics preserving (BoAP) library. The algorithm is implemented in two steps: 1) The image is decomposed into multiple resolutions, 2) extraction of bag-of-aesthetics features. The HSV colour space, local binary patterns, and saliency map extract features from the images. The AdaBoost classifier is trained and tested on a dataset of 3k images downloaded from DPChallange~\cite{27} and Flickr~\cite{91}, providing 92.06\% accuracy. Figure~\ref{fig:content_methods}(c) shows the framework of the BoAP algorithm.

\item To evaluate the quality of Chinese handwriting, Rongju~\etal~\cite{ref39} explored the problem of artificial intelligence with aesthetic feature representation. The first step extracts component layout features and global features from Chinese handwritten images. For components, the semi-automatic component extraction method extracts layout feature strokes. Similarly, the alignment, stability, and distribution of white spaces and gaps between strokes are global features extracted from input Chinese handwritten images. A novel dataset named  Chinese Handwriting Aesthetic Evaluation Database (CHAED)~\cite{128}  is built and used to train the SVM classifier. Finally, neural networks are trained on CHAED to get the aesthetic evaluation ability.

\end{enumerate}

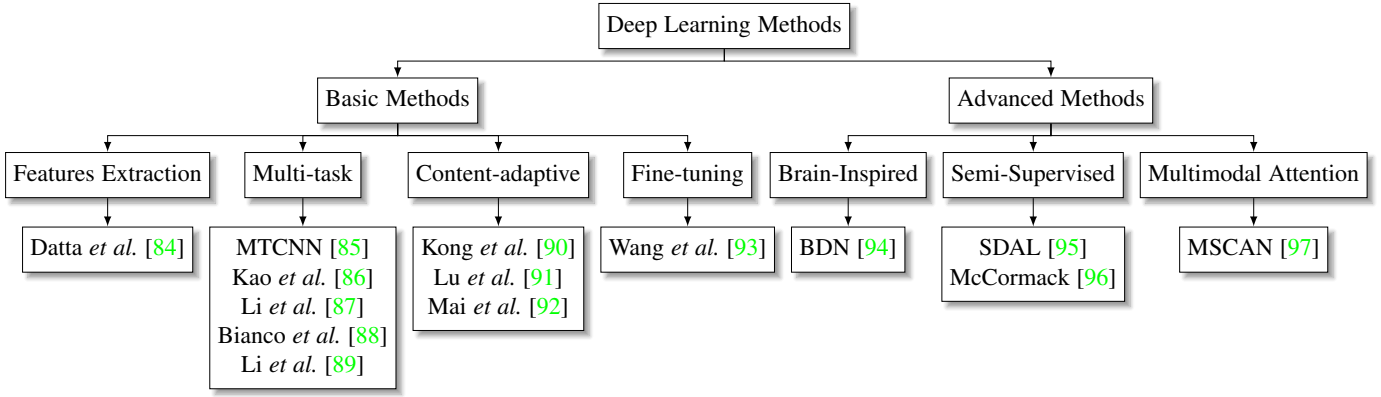
\begin{figure*}[t]
\centering
\resizebox{\textwidth}{!}{
\begin{forest}
forked edges,
for tree={draw,align=center,edge={-latex},fill=white,blur shadow}
[Deep Learning Methods
 [Basic Methods
    [Features Extraction
        [Datta~\etal~\cite{datta2006studying}]
    ]
    [Multi-task
        [MTCNN~\cite{129}\\
        Kao~\etal~\cite{ref17}\\
        Li~\etal~\cite{ref27a}\\
        Bianco~\etal~\cite{ref18}\\
        Li~\etal~\cite{ref29a}]
    ]
    [Content-adaptive
        [Kong~\etal~\cite{ref19}\\
        Lu~\etal~\cite{ref20}\\
        Mai~\etal~\cite{ref21}]
    ]
    [Fine-tuning
        [Wang~\etal~\cite{ref22}]
    ]
  ]
 [Advanced Methods
    [Brain-Inspired
        [BDN~\cite{ref24}]
    ]
    [Semi-Supervised
        [SDAL~\cite{ref25a}\\
        McCormack~\cite{ref30a}]
    ]
    [Multimodal Attention
       [ MSCAN~\cite{ref31a}]
    ]
 ]
]
\end{forest}}
\vspace{1mm}
	\caption{Overview of the deep learning methods and their classification based on similarity in structure}
	\label{fig:deeptaxanomy}
\end{figure*}

\section{Deep Learning Methods}
Deep learning uses artificial neural networks to automatically learn complex low and high-level features useful for computer vision tasks. In many cases, deep learning has produced results compared to human accuracy or even surpassed humans in many areas. Convolutional neural networks are the backbone of deep learning for image analysis \cite{khan2019dimension,khan2019person}. Once trained on millions of images, these networks can provide outstanding accuracy on image understanding tasks such as image aesthetic assessment. In this section, we discuss various important works for image aesthetics prediction using deep learning methods, shown in Figure~\ref{fig:deeptaxanomy}.

The aesthetic quality assessment of photographs can be formulated as a classification or regression or the combination of classification and regression approaches. There is a lack of consensus on the definition of aesthetic quality as it is a subjective matter. However, the photo-sharing communities rated the photos, and the average score is usually taken as the quality of the images and used as ground truth for different algorithms. Therefore, the quality of the assessment task is taken as a classification problem. Nevertheless, the problem can also be formulated as a regression to regress the quality of photographs to aesthetic score. Thus, the aesthetic quality assessment feature could be either extracted as a hand-crafted or learned using deep learning architecture in multi-task settings. The multi-task approaches tend to learn better and improve the aesthetic score significantly. 

\subsection{Deep learning Basic Methods}
\subsubsection{Deep Features Extraction Based Methods}
\begin{figure}[t]
	\centering
	\includegraphics[width=\columnwidth]{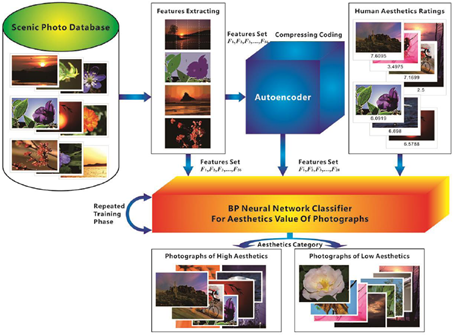}
	\caption{The scheme for deep features extraction methods~\cite{ref15}.}
	\label{fig3}
\end{figure}

The author used a deep neural network and extracted 56 visual features originally proposed by Datta~\etal~\cite{datta2006studying} for aesthetic assessment~\cite{ref15}. The dataset is collected from the internet consisting of 28896 images, where each image is resized to 160$\times$120 resolution, and features are extracted from images by converting them to HSV colour space. These extracted features include brightness, Earth Mover Distance (EMD), Hue, saturation, \etc The autoencoder has been used to compress the raw features into new features with $1/2$ or $1/4$ of its original input. The Artificial Neural Network (ANN) trains the network with both the extracted 56 visual features and the new features extracted from the autoencoder. The ANN is used here as a classifier to classify the photograph into two aesthetic categories: high and low aesthetics. Moreover, Convolutional Neural Networks and Deep Belief networks are also used for aesthetic evaluation on a larger dataset. The overall structure of their proposed scheme is depicted in Figure~\ref{fig3}. This scheme is tested on an AMD Athlon II PC providing 82.1\% accuracy. A global, local, and scene-aware information of images are considered and exploited the composite features extracted from corresponding pre-trained deep learning models for classification using SVM~\cite{ref26a}. They found that a deep residual network could produce more aesthetics-aware image representation and composite features. 

\subsubsection{Multi-task Convolutional Networks}
\begin{figure*}[t]
\centering
\begin{tabular}{cc} 

 \includegraphics[width=\columnwidth]{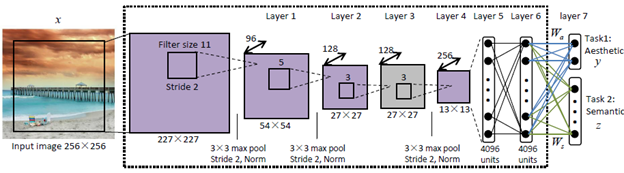}& 
\raisebox{22mm}{\multirow{4}{*}{\includegraphics[angle=270, width=0.8\columnwidth]{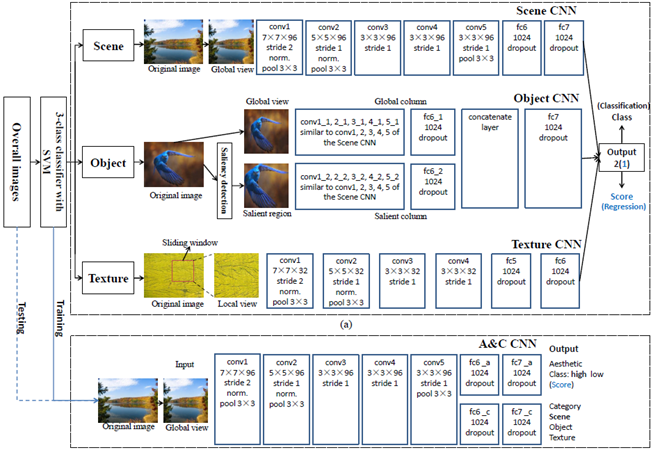}}}\\
 a)&  \\
\includegraphics[width=\columnwidth]{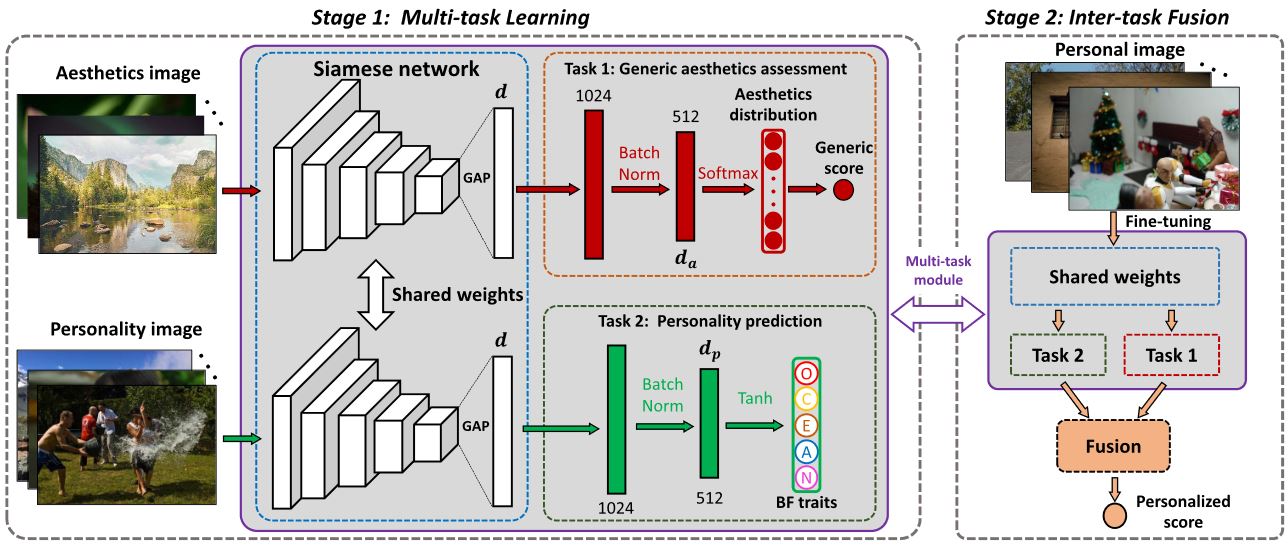}& \\
 c) & \\
 \includegraphics[width=\columnwidth]{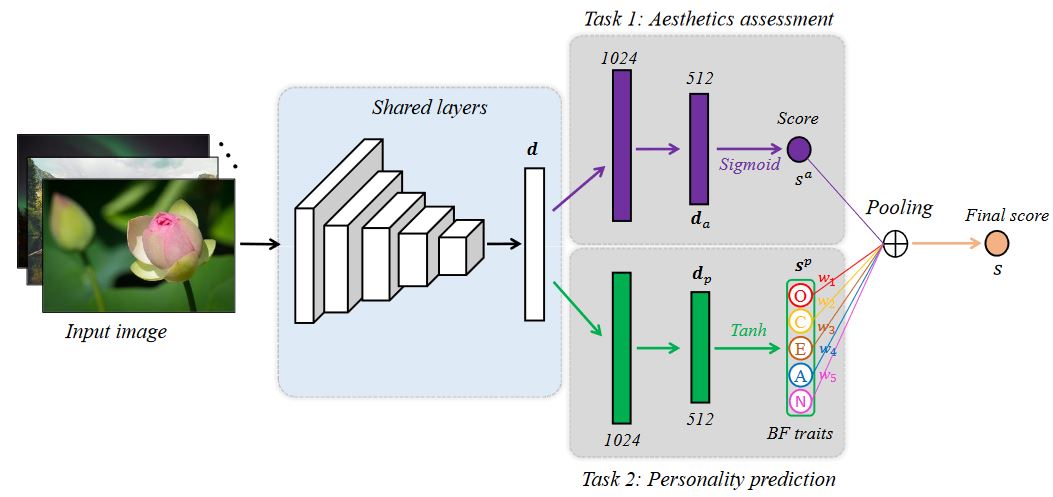}&\\ 
d) & b)\\

\end{tabular}
    \caption{Multi-task Convolutional Networks Based Methods. 	a) Architecture of the system proposed by Kao~\etal~\cite{ref16}, b) Framework of the system proposed by Kao~\etal~\cite{ref17},	c) The personality-assisted multi-task learning model by Li~\etal~\cite{ref29a}, and d) The architecture of the multi-task learning model by Li~\etal~\cite{ref27a}.}
    \label{fig:multi_task}
\end{figure*}

A Multi-task learning approach is employed to explore the correlation between automatic aesthetic assessment and semantic information. The task is to utilize semantic information in the joint objective function to improve the quality assessment task~\cite{ref16}. The approach provides both aesthetic and semantic labels as output. A Multi-Task Convolutional Neural network (MTCNN)~\cite{129} is designed that performs both semantic recognition and quality assessment considering an input image size of 227$\times$227. The proposed CNN automatically learns the relation between semantics and aesthetics. Their CNN consists of five convolutional layers, three pooling layers, and three fully-connected layers. The proposed Convolutional Neural Network architecture is shown in Figure~\ref{fig:multi_task}(a). Furthermore, three representations of the MTCNN are proposed in which different configurations of convolutional layers and pooling layers \cite{ref79, ref80, ref81} are designed. A multi-task probabilistic framework is applied. The network is trained and tested on the AVA dataset~\cite{murray2012ava} and Photo.net~\cite{ref3} dataset. AVA dataset consists of 255k images, and the photo.net dataset comprises 20,278 images. On the AVA dataset, MTCNN achieves up to 77.71\% accuracy, and on Photo.net, it achieves up to 65.20\% accuracy.

A Convolutional Neural Network-based framework has been proposed for the visual quality assessment~\cite{ref17}. There are three categories defined for each image; scene, object, and texture. Firstly, each image is classified into one of the three categories using SVM. Then for each category, a separate convolutional neural network named Scene CNN, Object CNN, and Texture CNN is trained to learn features and classify the output into a high aesthetic or low aesthetic class and a numerical aesthetic score. In addition, another single CNN called A\&C CNN is deployed, which performs recognition of quality and aesthetic ratings simultaneously for overall images. Figure~\ref{fig:multi_task}(b) shows the overall structure of the implemented scheme. The algorithm is tested on an AVA dataset containing 255k images. It achieves 91.3\% accuracy. The scene, object, and texture CNN are highly dependent on the classification accuracy of the SVM classifier. If SVM provides the wrong classification, the incorrect CNN gets activated and outputs inaccurate results.

An end-to-end personality-driven multi-task deep learning model has been introduced to assess the aesthetics of an image~\cite{ref27a} as shown in Figure~\ref{fig:multi_task}(c). Firstly, image aesthetics and personality traits are learned from the multi-task model. Then the personality features are used to modulate the aesthetics features, producing the optimal generic image aesthetics scores. 

Bianco~\etal~\cite{ref18} used deep learning to predict image aesthetics using aesthetic visual analysis (AVA)~\cite{murray2012ava} dataset. This model fine-tuned canonical convolutional neural network architecture to obtain aesthetic scores in this model. Aesthetic quality assessment is treated as a regression problem. Caffe network~\cite{ref82} is selected to be fine-tuned, and the last fully connected layer of CaffeNet is replaced by a single neuron providing an aesthetic score between 1 and 10. Another modification is incorporated in Caffe Net to use Euclidean loss~\cite{ref83} instead of Softmax loss~\cite{ref84}. A stochastic gradient descent backpropagation algorithm fine-tunes the new network. The dataset contains 255k images, from which 250,129 images are used for training and 4970 images for testing. The algorithm achieves 83\% accuracy.

A personality-assisted multi-task deep learning framework is presented~\cite{ref29a} as shown in Figure~\ref{fig:multi_task}(d) for both generic and personalized image aesthetics assessment. Initially, they introduced a multi-task learning network with shared weights to predict the aesthetics distribution of an image and Big-Five (BF) personality traits of people who like the image. They then used an inter-task fusion to generate individuals' personalized aesthetic scores on the image. 

\begin{figure*}[t]
\centering
\begin{tabular}{cc} 
\includegraphics[width=\columnwidth]{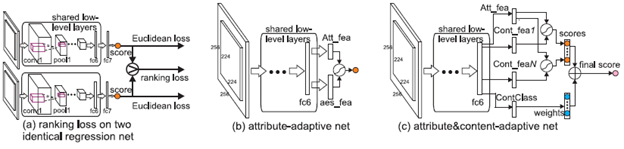}&
\multirow{5}{*}{\includegraphics[width=\columnwidth]{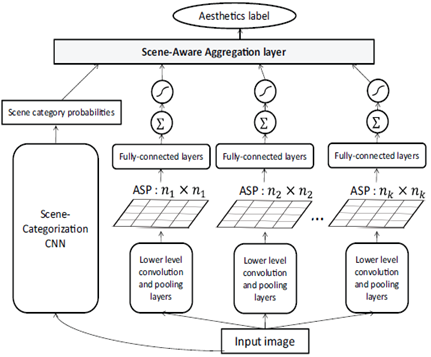}} \\
a) &\\
\includegraphics[width=\columnwidth]{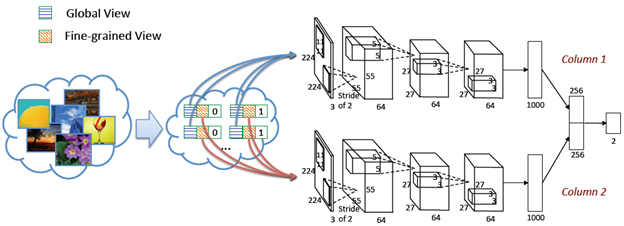}&\\
 b) &\\

\raisebox{2mm}{\includegraphics[width=\columnwidth]{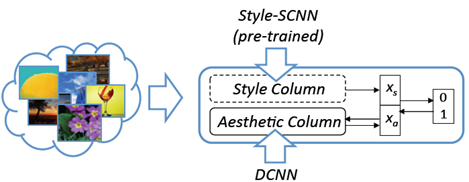}}& \\
c) & d) \\
\end{tabular}
    \caption{Content-adaptive Deep Learning Methods. a) Architectures of different models proposed by Kong~\etal~\cite{ref19}, b) Double column convolutional neural network model for aesthetic quality assessment proposed by Lu~\etal~\cite{ref20}, c) Regularized double-column convolutional neural network model proposed by Lu~\etal~\cite{ref20}, and d) Scene aware multi-net regression model of Mai~\etal~\cite{ref21}.}
    \label{fig:Content_adaptive}
\end{figure*}

\subsubsection{Content-Adaptive Deep Learning Methods}
A content adaptation technique using deep CNN has been proposed for image quality aesthetic assessment~\cite{ref19}. A new dataset is published by these researchers, which they named as Aesthetics and Attributes Database (AADB)~\cite{130} comprising 10k images. AlexNet architecture~\cite{ref88} is fine-tuned on AADB dataset. Softmax loss is replaced by Euclidean loss. Another Siamese network \cite{ref89, ref90} is fine-tuned with content category classification and attribute layers to achieve hybrid performance. An attribute-adaptive model and a content-adaptive model are designed. Figure~\ref{fig:Content_adaptive}(a) shows three different models initially based on AlexNet. Model (a) uses shared low-level layers of AlexNet and adopts Euclidean loss and Ranking loss, whereas model (b) is an attribute-adaptive net with an additional attribute predictor branch. Model (c) provides a combined adaptive net and attribute adaptive net approach. It takes an input image of size 227$\times$227 and provides 77.33\% accuracy.

A two-column content-adaptive aesthetic rating neural network is proposed that takes into account both style contents and semantic information~\cite{ref20}. Each column is trained on two different crops of a single image. Each column consists of three convolutional layers and three pooling layers followed by a fully connected layer. Finally, style and semantic features extracted by both columns are fused by two fully connected layers as shown in Figure~\ref{fig:Content_adaptive}(b). The network is trained using end-to-end learning and stochastic gradient descent. A network adaptation strategy is proposed to facilitate content-based image aesthetics. This helps improve the adaptation of images' semantic contents; hence, fewer images from each category are required for training. A Regularized Double-column Convolutional Neural Network (RDCNN) is proposed, which includes a single Style Column Convolutional Neural Network (Style-SCNN) for style information and a Double-Column Convolutional Neural Network (DCNN) for semantic information. The final structure of the framework is shown in Figure~\ref{fig:Content_adaptive}(c). This network is tested on the AVA dataset and IAD dataset~\cite{131} to categorize images into high and low quality and achieves 71.2\% accuracy.

A composition preserving convolutional neural network has been proposed for photo aesthetic assessment~\cite{ref21}. The network incorporates the concept of image quality degradation by resizing and clipping. Multi Net Adaptive spatial pooling Convolutional Neural Network (MNA-CNN) is designed to rate variable size images. For this purpose, an adaptive spatial pooling layer is introduced that adjusts its receptive size according to output rather than input. There are multiple streams of network~\cite{ullah2019two} where an adaptive spatial pooling layer replaces the last pooling layer. Pre-trained VGG~\cite{132} is fine-tuned on Torch Deep Learning package~\cite{133}, and each sub-network is trained separately. Another scene categorization CNN is trained on Places205-GoogleLeNet consisting of 2.5 million images. This framework is shown in Figure~\ref{fig:Content_adaptive}(d). Scene categorization network increases aesthetic assessment accuracy to 77.1\% accuracy.

\begin{figure}[t]
	\centering
	\includegraphics[width=\columnwidth]{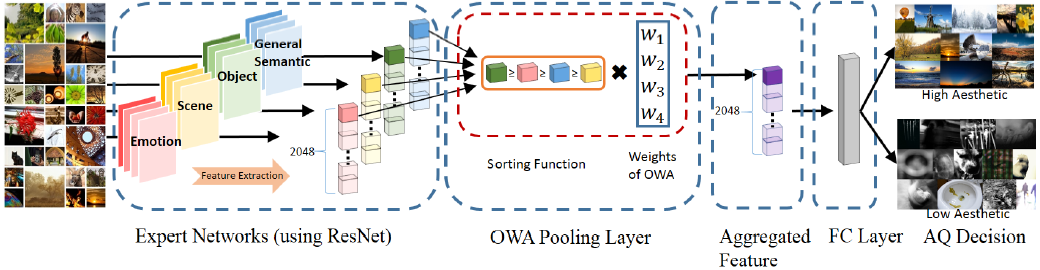}
	\caption{Deep semantic aggregation network proposed by Lu~\etal~\cite{ref23}.}
	\label{fig23}
\end{figure}

\subsubsection{Fine-tuning Based Approaches}
A pre-trained convolutional neural network is fine-tuned for assessing the quality of images~\cite{ref22}. AlexNet and VGG are fine-tuned to provide output in two categories (high and low). VGG is a deeper network than AlexNet, providing high accuracy and requiring more training time. AlexNet comprises five convolutional layers with ReLU non-linearity, five pooling layers, and three fully-connected layers. The last layer is replaced by a fully connected layer for a two-class classification. VGG is a deeper network consisting of sixteen to nineteen convolutional and pooling layers. Both global and local views train the networks. AVA and CUHKPQ datasets are used to fine-tune and are trained on both the global and local views. AlexNet achieves 91.20\% accuracy CUHKPQ dataset, and VGG achieves 91.93\% accuracy. AlexNet achieves 83.24\% accuracy on the AVA dataset, and VGG achieves 85.41\% accuracy.

A ResNet152 network has been used for image aesthetic quality assessment~\cite{ref23}, which was trained on the ImageNet dataset for object classification and further fine-tuned on AVA, Places, and emotion6 datasets. The network is trained for four different categories; scene images, object images, emotion images, and general semantic images as depicted in Figure~\ref{fig23}. For the scene images, 2.5 million images from the Places dataset~\cite{135} are used to fine-tune ResNet152. The network is trained using the AVA dataset for object images, and the emotion images network is trained on the Emotion6 dataset consisting of 1980 images. This network achieves 78.6\% accuracy.

\begin{figure*}[t]
\centering
\begin{tabular}{cc} 
\includegraphics[width=\columnwidth]{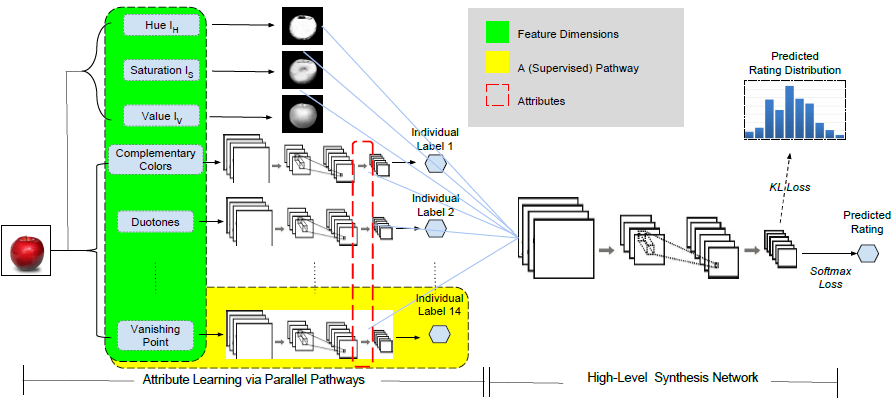}&
\includegraphics[width=\columnwidth]{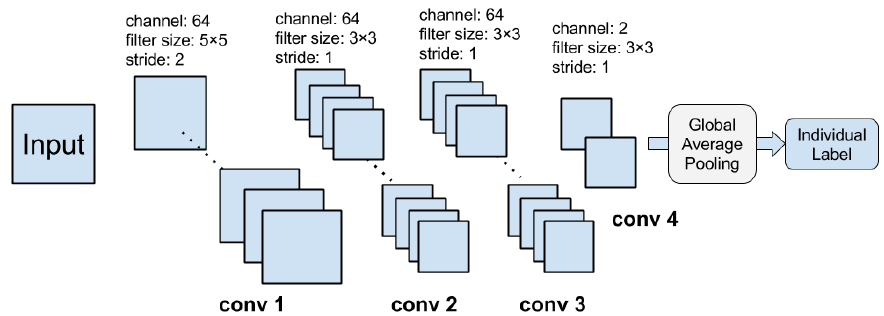}\\
a) & b) \\
\end{tabular}
\caption{Brain-inspired Approaches of~\cite{ref24}. a)~Brain inspired network architecture and b)~Attribute learning via parallel pathways.}
\label{fig:Brain_methods}
\end{figure*}

\subsection{Advanced Deep Methods}
\subsubsection{Brain-inspired Approaches} 
A Brain-inspired Deep Neural Network (BDN) has been proposed for image aesthetic assessment~\cite{ref24} and is composed of two parts. The first part is attribute learning via parallel pathways, and the second part is a high-level synthesis network as shown in Figure~\ref{fig:Brain_methods}(a). Attribute learning via parallel pathways is a combination of deep neural network streams. Different attributes are learned from input images, including hue, saturation, value, complementary colours, duotones, high dynamic range, image grain, light on white, long exposure, macro, motion blur, negative image, rule of thirds, shallow DOF, silhouettes, soft focus and vanishing point. Hue, saturation, and value are directly computed from the image, whereas the other attributes are learned using parallel deep neural networks as shown in Figure~\ref{fig:Brain_methods}(b). This network predicts a label 0 or 1  and is trained using the AVA dataset. Their high-level synthesis network is a four-layer convolutional neural network. This network predicts the overall aesthetic level of the image. At this stage, the entire network is trained end-to-end using the AVA dataset. Experiments are performed on 12 CPUs (Intel Xeon 2.7 GHz) and a GPU (Nvidia GTX680). Training and fine-tuning take around one day with an accuracy of 76.80\%.

\subsubsection{Semi-Supervised Approaches}
For image aesthetic quality assessment, Liu~\etal~\cite{ref25a} proposed a semi-supervised deep active learning (SDAL) algorithm, which discovers how humans perceive semantically significant regions from many images partially assigned with contaminated tags. 

An adaptive fractional dilated convolution is developed~\cite{ref28a}, which is aspect-ratio-embedded, composition-preserving and parameter-free. The fractional dilated kernel is adaptively constructed according to the image aspect ratios, where the interpolation of the nearest two integers dilated kernels are used to cope with the misalignment of fractional sampling. 

A convolutional neural network is used to investigate the relationship between image measures, such as complexity, and human aesthetic evaluation, using dimension reduction methods to visualize both genotype and phenotype space to support the exploration of new territory in a generative system~\cite{ref30a}. Convolutional neural networks trained on the artist's prior aesthetic evaluations are used to suggest new possibilities similar to or between known high-quality genotype-phenotype mappings. 

\subsubsection{Multimodal Attention-Based Networks}
\begin{figure}[t]
	\centering
	\includegraphics[width=\columnwidth]{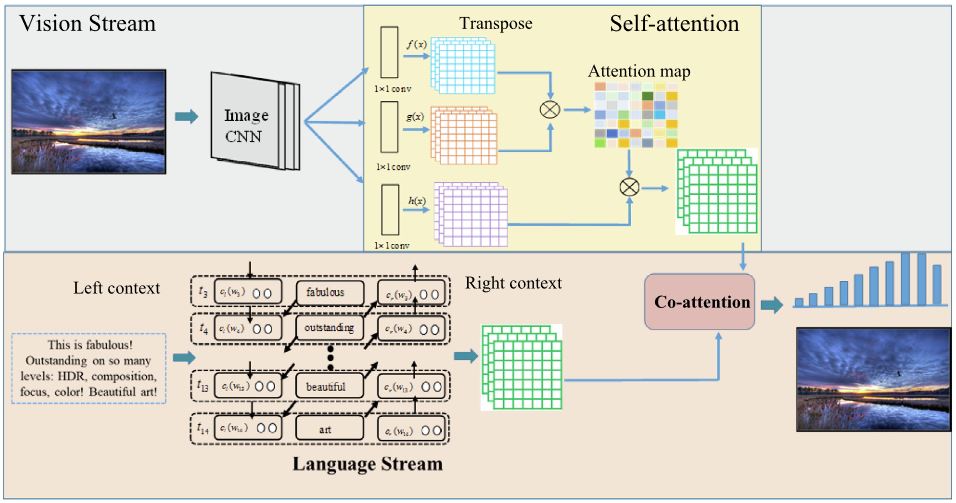}
	\caption{The multimodal self and collaborative attention network by Zhang~\etal~\cite{ref31a}.}
	\label{fig17b}
\end{figure}

The MSCAN, a multimodal self, and collaborative attention network is proposed for aesthetic prediction task~\cite{ref31a} as shown in Figure~\ref{fig17b}. The self-attention module finds the response at a position by attending to all positions in the images to encode spatial interaction of the visual elements. To model the complex image-textual feature relations, a co-attention module is used to perform textual-guided visual attention and visual-guided textual attention jointly. 

\section{Experimental Settings}
To make the survey more comprehensive, we first provide information about the publicly available widely used benchmark datasets and evaluation metrics, followed by the hand-crafted and deep learning comparisons.

\subsection{Datasets}
\subsubsection{Photo.net}
This dataset~\cite{ref3} is collected from \enquote{Photo.net}, a website of photo-sharing community established in 1997. The authors considered originality and aesthetic qualities used for rating photos on this website. Both the qualities are correlated, but originality is considered by the authors to be used for further processing due to its role in aesthetic value. The authors finally obtained 3581 photos for their work. The original dataset contains 20278 images.

\subsubsection{Aesthetic Visual Analysis Dataset}
Aesthetic Visual Analysis (AVA) dataset~\cite{murray2012ava} is derived from \enquote{dpchallenge.com} where the community uploads images to participate in different photographic challenges having titles and descriptions. In this connection, each image is linked with the information of its corresponding challenge that can provide the context of annotations when combined with aesthetic scores or semantic labels. 

AVA dataset contains 255,000 images that are associated with 963 challenges. While treating the aesthetic quality as a binary-class classification problem, images having an average aesthetic score value greater than the threshold value 5 + $\sigma$ are labelled as positive. In contrast, those with an average aesthetic score value less than 5 - $\sigma$ are negative. Training and testing sets contain 230,000 and 20,000 images respectively for a hard threshold $\sigma = 0$. Another split is also used to account for the top 10\% and bottom 10\% of the images, thus obtaining 25,000 images in the training set and 25,000 in the testing set.

\subsubsection{CUHK}
CUHK~\cite{Assessing2011Luca} is a publicly available dataset that contains photos of diversified aesthetic quality where 60,000 images were collected from \enquote{dpchallenge.com} each of which is rated by a minimum of 100 users. The images with top 10\% average rates are considered good category whereas the bottom 10\% average rates are considered bad category and, therefore, are manually examined. Due to the fact that CUHK draws a clear boundary between the classes, it is not a challenging dataset compared to the datasets where the class boundaries are not very clear.

\subsubsection{CUHK-PhotoQuality}
CUHK-PhotoQuality (CUHK-PQ) dataset \cite{luo2011content} is a collection of 17,690 images obtained from multiple online community platforms and university students. The images are aesthetically labelled either as high quality or low quality based on the feedback of independent viewers. The label for each image is decided only if eight reviewers out of ten favours it. CUHK-PQ dataset covers seven distinct categories: animal, plant, night, human, landscape, architecture, and static. The data is randomly partitioned according to 50-50 split to generate training and testing sets where the ratio of positive to negative samples is 1:3.

\subsubsection{MIRFLICKR}
In the domain of multimedia retrieval, MIRFLICKR dataset~\cite{murray2012ava} is a collection of 1 million images accompanied by textual tags, aesthetic annotations in the form of Flickr's interestingness, and EXIF metadata. As opposed to the AVA dataset, the MIRFLICKR dataset has an interestingness flag only that describes the aesthetic preference.   
Exposure and blur are two aspects associated with 44 visual concepts in the MIRFLICKR dataset. Images in this dataset are categorized in the following categories: neutral illumination, over-exposed, under-exposed, motion blur, no blur, out of focus, and partially blurred. 

\subsubsection{Aesthetics and Attributes Database }
Aesthetics and Attributes Database (AADB) dataset~\cite{130} is constructed by downloading 10k images from the Flickr website, where each image is rated by five raters independently. In this way, each image in the dataset is annotated with an aesthetic score and eleven attributes. The training, validation, and testing sets contain (8,500), (500), and (1,000) images, respectively. This dataset is distributed in different categories by the K-means clustering technique, where the value of k is set to ten based on experimental observation. 
 
\begin{table*}[t]
\setlength\extrarowheight{2pt}
\centering
\caption{Comparative analysis of handcrafted techniques using basic image features for image aesthetic assessment.}
\label{my-label}

\resizebox{\textwidth}{!}{
\begin{tabular}{l||c||cccccccccccc||ccc||cccccccc||c||c||c}\hline \hline
\multirow{2}{*}{Authors} & \multirow{2}{*}{Year} & \multicolumn{12}{c||}{Features}                                                                                                                      & \multicolumn{3}{|c||}{Classifiers}             & \multicolumn{8}{c||}{Dataset}                                                                         & \multirow{2}{*}{No. of Images} & \multicolumn{1}{c||}{Categorization Task} & \multirow{2}{*}{Accuracy} \\ \cline{3-25} \cline{27-27}
                        &                       & \rotatebox{90}{Color}   & \rotatebox{90}{Hue}      & \rotatebox{90}{Saturation} & \rotatebox{90}{Composition}    & \rotatebox{90}{Richness} & \rotatebox{90}{Contrast} & \rotatebox{90}{Brightness} & \rotatebox{90}{Simplicity} & \rotatebox{90}{Sharpness} & \rotatebox{90}{Texture}  & \rotatebox{90}{Depth \& Clarity} & \rotatebox{90}{Tone}     & \rotatebox{90}{SVM}      & \rotatebox{90}{Linear Regression} & \rotatebox{90}{Sparse Coding} & \rotatebox{90}{PhotoWeb} & \rotatebox{90}{DPChallenge} & \rotatebox{90}{Flicker}  & \rotatebox{90}{Photo Database} & \rotatebox{90}{CUHKPQ}   & \rotatebox{90}{CUHK}     & \rotatebox{90}{AVA}    & \rotatebox{90}{Self Collected} &                               & \rotatebox{90}{level}                        &                           \\ \hline
\rowcolor[HTML]{C0C0C0}  Ditta~\etal \cite{ref3}                                             & 2006                  & \checkmark & \checkmark & \checkmark   &                          &                                &            &                  &           &             &         &                 &         & \checkmark &                 &              & \checkmark &         &             &               &         &         &         &               & 3581                          & Bi                  & 71\%                                           \\
Li~\etal \cite{ref2}                                              & 2010                  & \checkmark &            &              &                          &                                & \checkmark & \checkmark       &           &             &         &                 &         &            & \checkmark      &              &         &            &             &               &         &         &         & \checkmark       & 510                        & Multi               & 72\%                                           \\ 
\rowcolor[HTML]{C0C0C0} Gadde~\etal \cite{ref5}                                             & 2011                  & \checkmark &            &              &                          &                                & \checkmark & \checkmark       &           &             &         &                 &         & \checkmark &                 &              &         & \checkmark &             &               &         &         &         &               & 12k                         & Bi                  & 79\%                                           \\ 
Pogacnik~\etal \cite{ref13}                                             & 2012                  & \checkmark &            &              & \checkmark               &                                &          &                    & \checkmark   &          &         &                 &         & \checkmark &                 &              &         & \checkmark & \checkmark  &               &         &         &         &               & 1306                          & Bi                  & 95\%                                           \\ 
\rowcolor[HTML]{C0C0C0} Lo~\etal \cite{ref1}                                              & 2013                  & \checkmark &            & \checkmark   & \checkmark               & \checkmark                     &          &                    &          &              &         &                 &         & \checkmark &                 &              &         &            &             & \checkmark    &         &         &         &               & 9651                          & Bi                  & 89\%                                           \\ 
Mavridaki~\etal~\cite{ref8}                                              & 2015                  & \checkmark &            &              & \checkmark               &                                &          &                    &           & \checkmark  &         &                 &         & \checkmark &                 &              &         &            &             &               & \checkmark & \checkmark & \checkmark &               & 12k                & Bi                  & 77.1\%                                        \\ 
\rowcolor[HTML]{C0C0C0} Redi~\etal~\cite{ref9}                                              & 2015                  & \checkmark &            &              &                          &                                & \checkmark &                  &           & \checkmark  & \checkmark &              &         & \checkmark &                 & \checkmark   &         &            &             &               &         &         & \checkmark &               & 250k                    & Bi                  & 75.7\%                                        \\ 
Aydin~\etal \cite{ref11}                                              & 2015                  & \checkmark &            &              &                          &                                &          &                    &           & \checkmark  &         & \checkmark      & \checkmark & \checkmark &              &              &         & \checkmark &             &               &         &         &         &               & 955                           & Bi                  &                    -                            \\ \hline \hline
\end{tabular}                                                                                                                         
}
\label{tab1}
\end{table*}
\begin{table*}[t]
\centering
\caption{Comparative analysis of hand-crafted techniques using texture and FG/BG features for image aesthetic assessment.}
\resizebox{\textwidth}{!}{

\begin{tabular}{l||c||cccccccccc||cc||ccccc||c||c||c}\hline \hline
\multirow{2}{*}{Methods} & \multirow{2}{*}{Year} & \multicolumn{10}{c||}{Features}                                                                                                                                                     & \multicolumn{2}{c||}{Classifiers} & \multicolumn{5}{c|}{Dataset}                                    & \multirow{2}{*}{Dataset Size} & \multicolumn{1}{c||}{Task} & \multirow{1}{*}{Accuracy} \\ \cline{3-19} \cline{21-22}
                         &                       & \rotatebox{90}{Color}    & \rotatebox{90}{\begin{tabular}[l]{@{}l@{}}Relative FG\\ Position\end{tabular}} & \rotatebox{90}{\begin{tabular}[l]{@{}l@{}}Visual Weight \\ Ration\end{tabular}} & \rotatebox{90}{Sentibank} & \rotatebox{90}{\begin{tabular}[l]{@{}l@{}}FG/ BG\end{tabular}} &\rotatebox{90}{ \begin{tabular}[l]{@{}l@{}}Spatio-Temporal\\ Binary Patterns\end{tabular}} & \rotatebox{90}{\begin{tabular}[l]{@{}l@{}}Global\\ Texture/features\end{tabular}} & \rotatebox{90}{\begin{tabular}[l]{@{}l@{}}Edge\\ Composition\end{tabular}} & \rotatebox{90}{\begin{tabular}[l]{@{}l@{}}Layout\\ Composition \& \\ Color Palette\end{tabular}} & \rotatebox{90}{\begin{tabular}[l]{@{}l@{}}Salient \\ Regions\end{tabular}} & \rotatebox{90}{SVR}             & \rotatebox{90}{SVM}            & \rotatebox{90}{Internet} & \rotatebox{90}{Flickr}   & \rotatebox{90}{NHK Video Database} & \rotatebox{90}{CUHK}     & \rotatebox{90}{Photo.net} &                                        & \rotatebox{90}{Level}          &                           \\ \hline \hline
\rowcolor[HTML]{C0C0C0}Yang~\etal~\cite{ref4}                                              & 2015                  & \checkmark & \checkmark                                                          & \checkmark                                                           &           &                                                  &                                                                               &                                                                   &                                                            &                                                                                  &                                                            & \checkmark        &                & \checkmark &          &                    &          &           & 431                          &   Multi                                     & 84.8\%                      \\ 
Bhattacharya~\etal~\cite{ref14}                                             & 2010                  &          & \checkmark                                                          &                                                                    &           &                                                  &                                                                               &                                                                   &                                                            &                                                                                  &                                                            & \checkmark        &                &          & \checkmark &                    &          &           & 632                                     &  Bi                & 87.3\%                      \\ 
\rowcolor[HTML]{C0C0C0}Bhattacharya~\etal~\cite{ref34}                                              & 2013                  &          &                                                                   &                                                                    & \checkmark  & \checkmark                                         & \checkmark                                                                      &                                                                   &                                                            &                                                                                  &                                                            &                 & \checkmark       &          &          & \checkmark           &          &           & 1k                                               &   Bi               &        -               \\ 
Lo~\etal~\cite{ref35}                                              & 2012                  & \checkmark &                                                                   &                                                                    &           &                                                  &                                                                               & \checkmark                                                          & \checkmark                                                   & \checkmark                                                                         &                                                            &                 & \checkmark       &          &          &                    & \checkmark &           &                                                 &    Bi              & 86.0\%                      \\ 
\rowcolor[HTML]{C0C0C0}Wang~\etal~\cite{ref41}                                              & 2010                  &          &                                                                   &                                                                    &           & \checkmark                                          &                                                                               & \checkmark                                                          &                                                            &                                                                                  & \checkmark                                                   &                 & \checkmark       &          &          &                    &          & \checkmark  & 3161                                                &     Bi             & 83.7\%                      \\ \hline \hline
\end{tabular}
}
\label{tab2}
\end{table*}

\begin{table*}[t]
\centering
\caption{Comparative analysis of hand-crafted techniques using local and global features for Bi-level image aesthetic assessment categorization task.}
\resizebox{\textwidth}{!}{%
\begin{tabular}{l||c||cccccc||ccccccc||c||c||c}\hline \hline
\multirow{2}{*}{Authors} & \multirow{2}{*}{Year} & \multicolumn{6}{c||}{Features}              & \multicolumn{7}{c||}{Classifiers}                   & \multirow{2}{*}{Dataset} & \multirow{2}{*}{No. of Images} & \multirow{2}{*}{Accuracy} \\ \cline{3-15}
                         &                       & \rotatebox{90}{SIFT} & \rotatebox{90}{GIST} & \rotatebox{90}{SDAM} & \rotatebox{90}{DCT}  & \rotatebox{90}{Wavelet} & \rotatebox{90}{SURF} & \rotatebox{90}{BN}   & \rotatebox{90}{SVR}  & \rotatebox{90}{KNN}  & \rotatebox{90}{Bayesian} & \rotatebox{90}{SVM}  & \rotatebox{90}{KLD}  & \rotatebox{90}{ANN}  &                          &                               &                           \\ \hline \hline
\rowcolor[HTML]{C0C0C0}  Gao~\etal~\cite{ref7}                     & 2015                  & \checkmark  & \checkmark  &      &      &         &      & \checkmark  & \checkmark  &      &          &      &      &      & Proprietary              & 2222                          & 72.7\%                    \\ 
Yin~\etal~\cite{ref10}                       & 2012                  &      & \checkmark  & \checkmark  &      &         &      &      &      & \checkmark  &          & \checkmark  &      &      & Flicker                  & 10200                         & 73\%                      \\ 
\rowcolor[HTML]{C0C0C0}  Saad~\etal~\cite{ref43}                    & 2012                  &      &      &      & \checkmark  &         &      &      &      &      & \checkmark      &      &      &      & Live                     & 799                           & 91\%                      \\ 
Wang~\etal~\cite{ref44}                        & 2005                  &      &      &      &      & \checkmark     &      &      &      &      &          &      & \checkmark  &      & Live                     & 489                           & 92\%                      \\ 

\rowcolor[HTML]{C0C0C0}   Riaz~\etal~\cite{ref45}                       & 2012                  &      &      &      &      & \checkmark     & \checkmark  &      &      &      &          &      &      & \checkmark  & Photo.net                & 250                           & 83\%                      \\ \hline \hline
\end{tabular}
}
\label{tab3}
\end{table*}
\subsection{Evaluation Metrics}
The most commonly used evaluation metrics in image aesthetic assessment are summarized in subsections.

\subsubsection{Overall Accuracy}
Overall accuracy (OA) takes into account True Positive (TP), True Negative (TN), False Positive (FP), and False Negative (FN) samples of a dataset. Accuracy may be misleading in the case of imbalanced data. However, it is a widely used measure to assess the performance of a classification model. It can be expressed mathematically as 

\begin{equation}
OA =  \frac{(TP+TN)}{(TP+FN+TN+FP)}\times 100
\label{Equation_accuracy}
\end{equation}

\subsubsection{Balanced Accuracy}
In the case of an imbalanced dataset, \emph{Balance Accuracy} (BA) can be used to evaluate the performance of a classifier and averaging recall values can calculate it for each class. Balance accuracy is computed as the arithmetic mean of sensitivity and specificity. Mathematically, it can be expressed by Eq.~\eqref{Equation_BalancedAccuracy}.

\begin{equation}
BA= \frac{Sensitivity + Specificity}{2}
\label{Equation_BalancedAccuracy}
\end{equation}

\emph{Sensitivity} is the true positive rate that computes the correctly predicted positive samples out of total positive samples, whereas specificity is the true negative rate that computes the correctly predicted negative samples out of total negative samples. Sensitivity and specificity are given below

\begin{equation}
Sensitivity= \frac{TP}{TP + FN}
\label{Equation_sensitivity}
\end{equation}
\begin{equation}
Specificity = \frac{TN}{TN + FP}
\label{Equation_specificity}
\end{equation}

\subsubsection{Precision-recall curve}
When the classes are highly imbalanced, the precision-recall curve is beneficial in assessing the performance of a classification model. The precision-recall curve highlights the trade-off between precision and recall for various threshold values. Higher the value of area under the precision-recall curve, higher are the values of recall and precision where the high precision value indicates a low false-positive rate and high recall value shows a low false-negative rate
\begin{equation}
Precision = \frac{TP}{TP + FP}
\label{Equation_precision}
\end{equation}
\begin{equation}
Recall = \frac{TP}{(TP + FN},
\label{Equation_recall}
\end{equation}
where precision and recall are given in Eq.~\eqref{Equation_precision}~and~\eqref{Equation_recall}, respectively. Here, precision refers to the number of true positives over the number of true positives and the number of false positives predicted by the classifier. On the other hand, recall indicates the number of true positives over the total number of positives, including true positives and false negatives in the positive class.
\begin{table*}[t]
\centering
\caption{Comparative analysis of hand-crafted techniques using content based features for Bi-level image aesthetic assessment categorization task.}
\resizebox{\textwidth}{!}{%
\begin{tabular}{l||c||cccccc||ccc||ccccccc||c||c}\hline \hline
\multirow{2}{*}{Authors} & \multirow{2}{*}{Year} & \multicolumn{6}{c||}{Features}                                                                               & \multicolumn{3}{c||}{Classifiers} & \multicolumn{7}{c||}{Dataset}                                     & \multirow{2}{*}{No. of Images} & \multirow{2}{*}{Accuracy} \\ \cline{3-18}
                         &                       & \rotatebox{90}{Bag-of-Words} & \rotatebox{90}{Fish Vector} & \rotatebox{90}{Dark Channel } & \rotatebox{90}{SCL}  & \rotatebox{90}{Bag-of-Aesthetics} & \rotatebox{90}{Global Features} & \rotatebox{90}{SVM}      & \rotatebox{90}{Adaboost}    & \rotatebox{90}{GMM}     & \rotatebox{90}{DPChallenge} & \rotatebox{90}{Photo.net} & \rotatebox{90}{CHUK} & \rotatebox{90}{CUHKPQ} & \rotatebox{90}{AVA}  & \rotatebox{90}{Flicker} & \rotatebox{90}{CHAED} &                               &                           \\ \hline \hline
\rowcolor[HTML]{C0C0C0}Nishiyama~\etal~\cite{ref6}                        & 2012                  & \checkmark          &             &                                   &      &                   &                 & \checkmark      &             &         & \checkmark         &           &      &        &      &         &       & 124664                        & 77.6\%                    \\ 
Marchesotti~\etal~\cite{ref12}                        & 2011                  & \checkmark          & \checkmark         &                                   &      &                   &                 & \checkmark      &             &         &             & \checkmark       & \checkmark  &        &      &         &       & 15581                         & 78.0\%                      \\ 
\rowcolor[HTML]{C0C0C0}Tang~\etal~\cite{ref36}                        & 2013                  &              &             & \checkmark                               &      &                   &                 & \checkmark      &             &         &             &           &      & \checkmark    &      &         &       & 17673                         & 83.0\%                      \\ 
Zhang~\etal~\cite{ref37}                       & 2014                  &              &             &                                   & \checkmark  &                   &                 &          &             & \checkmark     &             & \checkmark       & \checkmark  &        & \checkmark  &         &       & 40581                         & 85.5\%                    \\ 
\rowcolor[HTML]{C0C0C0}Su \etal \cite{ref38}                        & 2011                  &              &             &                                   &      & \checkmark               &                 &          & \checkmark         &         & \checkmark         &           &      &        &      & \checkmark     &       & 3k                          & 92.1\%                   \\ 
Sun \etal \cite{ref39}                         & 2015                  &              &             &                                   &      &                   & \checkmark             & \checkmark      &             &         &             &           &      &        &      &         & \checkmark   & -                             & -                         \\ \hline \hline
\end{tabular}
}
\label{tab4}
\end{table*}
\begin{table*}[t]
\centering
\caption{Comparative analysis of deep learning techniques for image aesthetic assessment.}
\resizebox{\textwidth}{!}{%
\begin{tabular}{l||c||ccc||c||ccccccccc||c||ccc||c}
\hline\hline
\multirow{2}{*}{Authors} & \multirow{2}{*}{Year} & \multicolumn{3}{c||}{Layers} & \multicolumn{1}{c||}{Models}                                                                           & \multicolumn{9}{c||}{Dataset}                                                               & \multirow{2}{*}{No. of Images}                                     & \multicolumn{3}{c||}{\begin{tabular}[c]{@{}c@{}}Classification\\ Level\end{tabular}} & \multirow{2}{*}{Accuracy} \\ \cline{3-15} \cline{17-19}
                         &                       & \rotatebox{90}{Convolutional} & \rotatebox{90}{Pooling} & \rotatebox{90}{Fully Connected} & \rotatebox{90}{\begin{tabular}[c]{@{}c@{}}Learning Model\\$\&$ Backbone\end{tabular}}                                                               & \rotatebox{90}{Dattra} & \rotatebox{90}{Photo.net} & \rotatebox{90}{ATA}  & \rotatebox{90}{\begin{tabular}[c]{@{}c@{}}Places205\\GoogleLeNet\end{tabular}} & \rotatebox{90}{AVA}  & \rotatebox{90}{CUHKPQ} & \rotatebox{90}{CHUK} & \rotatebox{90}{LIVE-IQ} & \rotatebox{90}{Lomas} &                                                                   & \rotatebox{90}{Bilevel}                   &\rotatebox{90}{Multilevel}                  & \rotatebox{90}{High-Low}                  &                           \\ \hline\hline
\rowcolor[HTML]{C0C0C0} Lu~\etal~\cite{ref20}                        & 2014                  & 6    & 6    & 3         &                                                                              &        &           &      &                       & \checkmark  &        &      &         &       & 255k                                                            & \checkmark                       &                             &                           & 71.20\%                    \\ 
Zhou~\etal~\cite{ref15}                       & 2015                  & 2    & 2    & 1         &                                                                              & \checkmark    &           &      &                       &      &        &      &         &       & 29k                                                             & \checkmark                       &                             &                           & 82.10\%                    \\ 
\rowcolor[HTML]{C0C0C0} Kao~\etal~\cite{ref16}                       & 2016                  & 5    & 3    & 1         &                                                                              &        & \checkmark       & \checkmark  &                       &      &        &      &         &       & 275k                                                            & \checkmark                       &                             &                           & 79.08\%                   \\ 
Mai \etal \cite{ref21}                       & 2016                  & 12   & 5    & 3         &                                                                              &        &           &      & \checkmark                   & \checkmark  &        &      &         &       & 255k                                                           & \checkmark                       &                             &                           & 77.10\%                    \\ 
\rowcolor[HTML]{C0C0C0} Wang~\etal~\cite{ref22}                        & 2016                  & 5    & 5    & 3         &                                                                              &        &           &      &                       & \checkmark  & \checkmark    &      &         &       & 273k                                                           &                           &                             & \checkmark                       & 91.93\%                   \\ 
Liu~\etal~\cite{ref25a}                        & 2018                  &      &      &           & \begin{tabular}[c]{@{}c@{}}Semi-Supervised\\ \& Active Learning\end{tabular} &        & \checkmark       &      &                       & \checkmark  &        & \checkmark  & \checkmark     &       & \begin{tabular}[c]{@{}c@{}}12k, 3581 \& \\ 250k, 779\end{tabular} & \checkmark                       &                             &                           & 94.65\%                   \\ 
\rowcolor[HTML]{C0C0C0} Fu \etal \cite{ref26a}                        & 2018                  &      &      &           & \begin{tabular}[c]{@{}c@{}}Different deep\\ \& Learning models\end{tabular}  &        &           &      &                       & \checkmark  &        &      &         &       & 250k                                                            &                           &                             & \checkmark                       & 90.01\%                   \\ 
Li~\etal~\cite{ref27a}                        & 2019                  &      &      &           & DenseNet121                                                                  &        &           &      &                       & \checkmark  &        &      &         &       & 250k                                                            & \checkmark                       &                             &                           & 81.50\%                   \\ 
\rowcolor[HTML]{C0C0C0} Chen~\etal~\cite{ref28a}                       & 2020                  &      &      &           & ResNet-50                                                                    &        &           &      &                       & \checkmark  &        &      &         &       & 250k                                                            & \checkmark                       &                             &                           & 83.24\%                   \\ 
Li~\etal~\cite{ref29a}                       & 2020                  &      &      &           & Siamese Network                                                              &        &           &      &                       & \checkmark  &        &      &         &       & 250k                                                            & \checkmark                       &                             &                           & 83.70\%                   \\ 
\rowcolor[HTML]{C0C0C0} McCormack $\&$ Lomas \etal \cite{ref30a}                       & 2021                  &      &      &           & ResNet-50                                                                    &        &           &      &                       &      &        &      &         & \checkmark   & 1774                                                              &                           & \checkmark                         &                           & 97.00\%                      \\ 

Zhang~\etal~\cite{ref31a}                       & 2021                  &      &      &           & InceptionNet                                                                 &        &           &      &                       & \checkmark  &        &      &         &       & 250k                                                            & \checkmark                       &                             &                           & 86.66\%                   \\ \hline\hline
\end{tabular}
}
\label{tab5}
\end{table*}

\subsection{Analysis}
This section provides a comparative performance analysis of both hand-crafted and deep learning-based methods. We provide an overview of how the reviewed techniques are different from each other with respect to features utilized, accuracy, dataset size, and classifiers used.

\subsubsection{Performance of Hand-Crafted Methods }
In this section, we show the analysis of various hand-crafted techniques as follows

\begin{itemize}
\item \textbf{Basic Feature Methods.} Table~\ref{tab1} presents the accuracy of basic feature methods results for image aesthetic assessment. Depending on the dataset used for testing, the accuracy varies significantly. The maximum accuracy is obtained by~\cite{ref13} for DPChallenge and Flicker datasets. However, recent methods such as ~\cite{ref9}~and~\cite{ref8} reports less accuracy as the datasets employed are different and the number of images is significantly higher. The basic feature method's popular choice for the classifier is SVM, and the essential feature is colour.

\item \textbf{Statistical Methods.} A comparison of accuracy, dataset size, classifier, and features extracted for statistical methods is given in Table~\ref{tab2}. In multi-level classification, Yang~\etal~\cite{ref4} achieves the highest accuracy of 84.83\% while in bi-level classification Lo~\etal~\cite{ref35} obtained 86\%. Although the authors employed differents for each classification level. SVM is mostly employed for classification.

\item \textbf{Global and Local Features Methods.}
The local and global features methods are provided in Table~\ref{tab3} showing the comparison of accuracy, datasets, the number of images, classifiers, year of publication, and attributes extracted. The accuracy for the mentioned methods ranges from 72.7\% to 92\%. The algorithms utilize different features, datasets, and classifiers for each technique.

\item \textbf{Content Based Methods.}
Table~\ref{tab4} gives the comparison between content-based hand-crafted methods. All the methods are evaluated for bi-level image aesthetic assessment tasks. The number of images employed by content-based is relatively higher than other previously mentioned methods. Most methods use SVM as a classifier while there is no set choice for features and datasets. It can also be observed that the higher the number of images in the dataset lower the accuracy and vice versa.

\end{itemize}

In summary, a large dataset is not required for hand-crafted methods. These techniques use a few hundred or a few thousand images to train classifiers. Almost 75\% of articles discussed in this survey utilized an SVM classifier to classify images into high and low aesthetic levels, and around 15\% used support vector regression. Here, the regression provides a continuous score on which threshold is applied for classification into different aesthetic levels. Hand-tuned approaches mainly rely on low-level features and do not consider semantic information of images, providing a minimal scoped aesthetic rating.

\subsubsection{Performance of Deep Learning Methods}
This section presents the comparative analysis of deep learning approaches in terms of layers, learning models, datasets, number of images per dataset, classification level, and accuracy. Table~\ref{tab5} shows the comparison of various deep learning techniques for image aesthetic assessment. Deep learning techniques provide better accuracy than hand-crafted techniques, focusing on the broader picture, including low-level and high-level features. The deep convolutional neural networks require considerable data for training. As the Table depicts, the datasets are more significant than those used in hand-crafted techniques. The depth of the network~\ie the number of layers for each method is also represented in Table~\ref{tab5}. Moreover, the accuracy may not be directly proportional to the depth of the network. One should also note that deep learning techniques require more computational resources and time for training and deployment. 

\section{Limitation and Challenges}
We here list some of the limitations and challenges in the following paragraphs.

\begin{itemize}
 \item \textbf{Lack of Dataset}: The algorithms are trained on various datasets; hence, there is no accurate way to determine the actual performance comparison. The best approach is to fix the dataset for training and evaluation.

 \item \textbf{Open Source Algorithms}: In image aesthetics, most of the algorithms and networks are not open source. The open-source codes are essential for future development and improvement.

 \item \textbf{Lack of Benchmark}: The image aesthetics lack a benchmark dataset to evaluate the algorithms, where each one reports the accuracy on the dataset of their choice. A standard benchmark will help accurately record algorithms' progress in image aesthetics. 

 \item \textbf{Parameters Comparison}: The methods in image aesthetics lack comparison on the number of parameters that are critical for many real-time computer vision applications. Unfortunately, existing models only focus on performance without giving any information about the number of parameters and efficiency, which may not be a true representation in the accuracy. Hence, attempts should be made for efficient models for deployment on real-time devices.  

 \item \textbf{Generalization}: is a challenging task, and many proposed models only work well on the suggested settings. The mentioned models perform better in one scenario due to their design for that specific task and fail in other settings. Further, the data can influence the generalization as well as robustness; thus, a significant step is to generalize these algorithms on more generalized tasks.

\end{itemize}

\section{Conclusion}
Images may be degraded due to compression artifacts, illumination or lighting issues, pose or camera angle, sensor problems, background clutter, and other imperfections. Image quality assessment can quantify such degradations, which can then be analyzed for corrections. With the rapid use of digital photographs in almost every field of life like medical, communication, safety, security, entertainment, sports \etc, image quality assessment becomes a basic needed functionality. This paper provides a detailed review of image aesthetic assessment techniques available in the literature, along with their strengths and shortcomings. The accuracy, dataset size, classifier selection, and the choice of deep learning models used for each technique are also presented. Our paper also provides an insight for future research to compare the performance of different image aesthetic assessment algorithms.
\balance
\bibliographystyle{ieeetr}
\bibliography{refs}

\begin{thebibliography}{100}

\bibitem{kong2016photo}
S.~Kong, X.~Shen, Z.~Lin, R.~Mech, and C.~Fowlkes, ``Photo aesthetics ranking
  network with attributes and content adaptation,'' in {\em European Conference
  on Computer Vision}, pp.~662--679, Springer, 2016.

\bibitem{ref1}
K.-Y. Lo, K.-H. Liu, and C.-S. Chen, ``Intelligent photographing interface with
  on-device aesthetic quality assessment,'' in {\em Asian Conference on
  Computer Vision}, 2012.

\bibitem{ref3}
R.~Datta, D.~Joshi, J.~Li, and J.~Z. Wang, ``Studying aesthetics in
  photographic images using a computational approach,'' in {\em ECCV'06
  Proceedings of the 9th Eurpeon Conference on Computer Vision}, 2006.

\bibitem{ref9}
M.~Redi, N.~Rasiwasia, G.~Aggarwal, and A.~Jaimes, ``The beauty of capturing
  faces: Rating the quality of digital portraits,'' in {\em 2015 11th IEEE
  International Conference and Workshops on Automatic Face and Gesture
  Recognition (FG)}, vol.~1, 2015.

\bibitem{ref11}
T.~O. Aydin, A.~Smolic, and M.~Gross, ``Automated aesthetic analysis of
  photographic images,'' {\em IEEE Transactions on Visualization and Computer
  Graphics}, vol.~21, no.~14777775, pp.~31--42, 2015.

\bibitem{ref8}
E.~Mavridaki and V.~Mezaris, ``A comprehensive aesthetic quality assessment
  method for natural images using basic rules of photography,'' in {\em IEEE
  International Conference on Image Processing}, 2015.

\bibitem{ref2}
C.~Li, A.~C. Loui, and T.~Chen, ``Towards aesthetics: a photo quality
  assessment and photo selection system,'' in {\em MM'10 Proceedings of the
  18th ACM International Conference on Multimedia}, 2010.

\bibitem{ref13}
D.~Pogacnik, R.~Ravnik, N.~Bovcon, and F.~Solina, ``Evaluating photo aesthetic
  using machine learning,'' in {\em Slovenian KDD Conference on Data Mining and
  Data Warehouses (SiKDD) 2012}, 2012.

\bibitem{ref4}
W.~Yang, Q.~Tao, and H.~Wu, ``Figure and landscape photo quality assessment
  based on visual aesthetics,'' {\em Journal of Information and Computational
  Science}, vol.~1, no.~1, pp.~2477--2486, 2015.

\bibitem{ref14}
S.~Bhattacharya, R.~Sukthankar, and M.~Shah, ``A framework for photo-quality
  assessment and enhancement based on visual aesthetics,'' in {\em MM '10
  Proceedings of the 18th ACM international conference on Multimedia},
  pp.~271--280, 2010.

\bibitem{ref34}
S.~Bhattacharya, B.~Nojavansghari, T.~Chen, D.~Liu, S.-F. Chang, and M.~Shah,
  ``Towards a comprehensive computational model for aesthetic assessment of
  videos,'' in {\em MM'13 Proceedings of the 21st Internationak Conference on
  Multimedia}, pp.~361--364, 2013.

\bibitem{ref35}
K.~Lo, K.~Liu, and C.~Chen, ``Assessment of photo aesthetics with efficiency,''
  in {\em International Conference on Pattern Recognition}, 2012.

\bibitem{ref41}
L.-K. Wang and K.-L. Low, ``Salieny-enhanced image aesthetics class
  prediction,'' in {\em IEEE Int. Conference on Image Processing}, 2010.

\bibitem{ref7}
Z.~Gao, S.~Wang, and Q.~Ji, ``Multiple aesthetic attribute assessment by
  exploiting relations among aesthetic attributes,'' in {\em ACM on
  International Conference on Multimedia Retrieval}, 2015.

\bibitem{ref43}
M.~A. Saad, A.~C. Bovik, and C.~Charrier, ``Blind image quality assessment. a
  natural scene statistics approach in the dct domain,'' in {\em Transactions
  on Image Processing}, 2012.

\bibitem{ref10}
W.~Yin, T.~Mei, and C.~W. Chen, ``Assessing photo quality with geo-context and
  crowd sourced photos,'' in {\em Proceedings of VCIP}, 2012.

\bibitem{ref44}
Z.~Wang and E.~P. Simoncelli, ``Reduced-reference image quality assessment
  using a wavelet-domain natural image statistic model,'' in {\em Proceedings
  of SPIE - The International Society for Optical Engineering}, vol.~5666,
  pp.~149--159, 2005.

\bibitem{ref45}
S.~Riaz, K.~H. Lee, and S.-W. Lee, ``Aesthetic score assessment based on
  generic features in digital photography,'' in {\em AUN/SEED-Net Regional
  Conf. on Inf. and Communication Technology}, pp.~76--79, 2012.

\bibitem{ref6}
M.~Nishiyama, T.~Okabe, I.~Sato, and Y.~Sato, ``Aesthetic quality
  classification of photographs based on color harmony,'' in {\em IEEE
  Conference on Computer Vision and Pattern Recognition}, 2011.

\bibitem{ref12}
L.~Marchesotti, F.~Perronnin, D.~Larlus, and G.~Csurka, ``Accessing the
  aesthetic quality of photographs using generic image descriptors,'' in {\em
  International Conference on Computer Vision}, 2011.

\bibitem{ref36}
X.~Tang, W.~Luo, and X.~Wang, ``Content-based photo quality assessment,'' {\em
  IEEE Trans. on Multimedia}, vol.~15, pp.~1930--1943, 2013.

\bibitem{ref37}
L.~Zhang, Y.~Gao, V.~Zhang, Q.~Tian, and R.~Zimmermann, ``Perception-guided
  multimodel feature fusion for photo aesthetic assessment,'' in {\em MM'14
  Proceedings of 22nd ACM Internaltional Conference on Multimedia},
  pp.~237--246, 2014.

\bibitem{ref38}
H.-H. Su, T.-W. Chen, C.-C. Kao, W.~H. Hsu, and S.-Y. Chien, ``Scenic photo
  quality assessment with bag of aesthetics-preserving features,'' in {\em
  MM'11 Proceedings of the 19th ACM International Conference on Multimedia},
  pp.~1213--1216, 2011.

\bibitem{ref39}
R.~Sun, Z.~Lian, Y.~Tang, and J.~Xiao, ``Aesthetic visual quality evaluation of
  chinese handwritings,'' in {\em International Joint Conference on Artificial
  Intelligence}, pp.~2510--2516, 2015.

\bibitem{ref30}
Y.~Ke, X.~Tang, and F.~Jing, ``The design of high-level features for photo
  quality assessment,'' in {\em IEEE Computer Society Conference on Computer
  Vision and Pattern Recognition}, 2006.

\bibitem{uzair2017non}
M.~Uzair, A.~Mahmood, A.~Mian, and C.~McDonald, ``Periocular region-based
  person identification in the visible, infrared and hyperspectral imagery,''
  {\em Neurocomputing}, vol.~149, pp.~854--867, 2015.

\bibitem{ref31}
A.~Chatterjee and O.~Vartanian, ``Neuroscience of aesthetics,'' {\em Annals of
  the New York Academy of Sciences}, vol.~1369, pp.~172--194, 2016.

\bibitem{ref32}
Y.~Deng, C.~C. Loy, and X.~Tang, ``Image aesthetic assessment: An experimental
  survey,'' {\em IEEE Signal Processing Magazine}, vol.~34, no.~4, pp.~80--106,
  2017.

\bibitem{ref33}
L.~Barrett, B.~Masquita, K.~Ochsner, and J.~Gross, ``The experience of
  emotion,'' {\em Annual Review of Psychology}, vol.~58, pp.~373--403, 2007.

\bibitem{ref25}
H.~Maitre, ``A review of image quality assessment methods with application to
  computational photography,'' in {\em SPIE Proceedings of Multispectral Image
  Processing and Analysis}, vol.~9811, 2015.

\bibitem{ref26}
A.~Joy and S.~K, ``Aesthetic quality classification of photographs: A
  literature survey,'' {\em International Journal of Computer Applications},
  vol.~108, no.~15, 2014.

\bibitem{ref27}
Marchesotti, L., Murray, N., and Perronnin, ``Discovering beautiful attributes
  for aesthetic image analysis,'' {\em International Journal of Computer
  Vision}, vol.~1, no.~21, 2014.

\bibitem{ref28}
X.~Lu, Z.~Lin, X.~Shen, R.~Mech, and J.~Z. Wang, ``Deep multi-patch aggregation
  network for image style, aesthetics and quality estimation,'' in {\em IEEE
  International Conference on Computer Vision (ICCV)}, pp.~990--998, 2015.

\bibitem{ref29}
R.~Datta, J.~Li, and J.~Z. Wang, ``Algorithemic inferencing of aesthetics and
  emotion in natural images: An exposition,'' in {\em 15th IEEE International
  Conference on Image Processing}, 2008.

\bibitem{deng2017image}
Y.~Deng, C.~C. Loy, and X.~Tang, ``Image aesthetic assessment: An experimental
  survey,'' {\em IEEE Signal Processing Magazine}, vol.~34, no.~4, pp.~80--106,
  2017.

\bibitem{mahmood2018multi}
A.~Mahmood, M.~Uzair, and S.~Al-Maadeed, ``Multi-order statistical descriptors
  for real-time face recognition and object classification,'' {\em IEEE
  Access}, vol.~6, pp.~12993--13004, 2018.

\bibitem{ullah2019internal}
H.~Ullah, M.~Uzair, A.~Mahmood, M.~Ullah, S.~D. Khan, and F.~A. Cheikh,
  ``Internal emotion classification using eeg signal with sparse discriminative
  ensemble,'' {\em IEEE Access}, vol.~7, pp.~40144--40153, 2019.

\bibitem{yan2013learning}
J.~Yan, S.~Lin, S.~Bing~Kang, and X.~Tang, ``Learning the change for automatic
  image cropping,'' in {\em Proceedings of the IEEE conference on computer
  vision and pattern recognition}, pp.~971--978, 2013.

\bibitem{uzair2015hyperspectral}
M.~Uzair, A.~Mahmood, and A.~Mian, ``Hyperspectral face recognition with
  spatiospectral information fusion and pls regression,'' {\em IEEE
  Transactions on Image Processing}, 2015.

\bibitem{ullah2019stacked}
M.~Ullah, H.~Ullah, S.~D. Khan, and F.~A. Cheikh, ``Stacked lstm network for
  human activity recognition using smartphone data,'' in {\em 2019 8th European
  Workshop on Visual Information Processing (EUVIP)}, pp.~175--180, IEEE, 2019.

\bibitem{ref48}
T.-F. Lee, P.-J. Chao, H.-M. Ting, L.~Chang, Y.-J. Huang, J.-M. Wu, H.-Y. Wang,
  M.-F. Horng, C.-M. Chang, J.-H. Lan, Y.-Y. Huang, F.-M. Fang, and S.~W.
  Leung, ``Using multivariate regression model with least absolute shrinkage
  and selection operator (lasso) to predict the incidence of xerostomia after
  intensity-modulated radiotheraphy for head and neck cancer,'' {\em PLoS One
  Journal}, vol.~9, 2014.

\bibitem{khan2019disam}
S.~D. Khan, H.~Ullah, M.~Uzair, M.~Ullah, R.~Ullah, and F.~A. Cheikh, ``Disam:
  Density independent and scale aware model for crowd counting and
  localization,'' in {\em 2019 IEEE International Conference on Image
  Processing (ICIP)}, pp.~4474--4478, IEEE, 2019.

\bibitem{27}
``Dpchallenge.'' \url{https://www.dpchallenge.com/}.
\newblock [Online; accessed 09-Jan-2018].

\bibitem{89}
J.~H. Elder and S.~W. Zucker, ``Local scale control for edge detection and blur
  estimation,'' {\em IEEE Transactions on pattern analysis and machine
  intelligence}, vol.~20, no.~7, pp.~699--716, 1998.

\bibitem{ref47}
H.~Bay, A.~Ess, and L.~V. Gool, ``Speeded-up robust features (surf),'' {\em
  Journal of Computer Vision and Image Understanding}, vol.~110, no.~3,
  pp.~346--359, 2008.

\bibitem{90}
F.~J. Bianchi, C.~Booij, and T.~Tscharntke, ``Sustainable pest regulation in
  agricultural landscapes: a review on landscape composition, biodiversity and
  natural pest control,'' {\em Proceedings of the Royal Society B: Biological
  Sciences}, vol.~273, no.~1595, pp.~1715--1727, 2006.

\bibitem{ref74}
L.~Mai, H.~Le, Y.~Niu, and F.~Liu, ``Rule of thirds detection from
  photographs,'' in {\em IEEE International Symposium on Multimedia}, 2011.

\bibitem{luo2011content}
W.~Luo, X.~Wang, and X.~Tang, ``Content-based photo quality assessment,'' in
  {\em International Conference on Computer Vision}, pp.~2206--2213, IEEE,
  2011.

\bibitem{murray2012ava}
N.~Murray, L.~Marchesotti, and F.~Perronnin, ``Ava: A large-scale database for
  aesthetic visual analysis,'' in {\em 2012 IEEE Conference on Computer Vision
  and Pattern Recognition}, pp.~2408--2415, IEEE, 2012.

\bibitem{ref49}
F.~A-Osaimai, M.~Bennamoun, and A.~Mian, ``Illumination normalization for color
  face images,'' {\em International Sympoium on Visual Computing ISVC},
  pp.~99--101, 2006.

\bibitem{ref50}
D.~Basak, S.~Pal, and D.~C. Patranabis, ``Supprot vector regression,'' {\em
  Journal of Neural Information Processing}, vol.~11, no.~10, 2007.

\bibitem{91}
T.~Rattenbury, N.~Good, and M.~Naaman, ``Towards automatic extraction of event
  and place semantics from flickr tags,'' in {\em Proceedings of the 30th
  annual international ACM SIGIR conference on Research and development in
  information retrieval}, pp.~103--110, ACM, 2007.

\bibitem{121}
D.~Borth, T.~Chen, R.~Ji, and S.-F. Chang, ``Sentibank: large-scale ontology
  and classifiers for detecting sentiment and emotions in visual content,'' in
  {\em Proceedings of the 21st ACM international conference on multimedia},
  pp.~459--460, 2013.

\bibitem{ullah2019single}
M.~Ullah, H.~Ullah, and F.~A. Cheikh, ``Single shot appearance model (ssam) for
  multi-target tracking,'' {\em Electronic Imaging}, vol.~2019, no.~7,
  pp.~466--1, 2019.

\bibitem{ullah2020multi}
H.~Ullah, I.~U. Islam, M.~Ullah, M.~Afaq, S.~D. Khan, and J.~Iqbal,
  ``Multi-feature-based crowd video modeling for visual event detection,'' {\em
  Multimedia Systems}, pp.~1--9, 2020.

\bibitem{ullah2018anomalous}
H.~Ullah, A.~B. Altamimi, M.~Uzair, and M.~Ullah, ``Anomalous entities
  detection and localization in pedestrian flows,'' {\em Neurocomputing},
  vol.~290, pp.~74--86, 2018.

\bibitem{ullah2017density}
H.~Ullah, M.~Uzair, M.~Ullah, A.~Khan, A.~Ahmad, and W.~Khan, ``Density
  independent hydrodynamics model for crowd coherency detection,'' {\em
  Neurocomputing}, vol.~242, pp.~28--39, 2017.

\bibitem{ullah2019hybrid}
H.~Ullah, M.~Ullah, and M.~Uzair, ``A hybrid social influence model for
  pedestrian motion segmentation,'' {\em Neural Computing and Applications},
  vol.~31, no.~11, pp.~7317--7333, 2019.

\bibitem{ref51}
G.~Ye, D.~Liu, I.-H. Jhou, and S.-F. Chang, ``Robust late fusion with rank
  minimization,'' in {\em IEEE Conference on Computer Vision and Pattern
  Recognition (CVPR)}, 2012.

\bibitem{122}
S.~Bhattacharya, B.~Nojavanasghari, T.~Chen, D.~Liu, S.-F. Chang, and M.~Shah,
  ``Towards a comprehensive computational model foraesthetic assessment of
  videos,'' in {\em Proceedings of the 21st ACM international conference on
  Multimedia}, pp.~361--364, ACM, 2013.

\bibitem{uzair2020bio}
M.~Uzair, R.~Brinkworth, and A.~Finn, ``Bio-inspired video enhancement for
  small moving target detection,'' {\em IEEE Transactions on Image Processing},
  vol.~30, pp.~1232--1244, 2021.

\bibitem{ref42}
L.~Itti, C.~Koch, and E.~Niebur, ``A model of saliency-based visual attention
  for rapid scene analysis,'' {\em IEEE Trans. on Pattern Analysis and Machine
  Intelligence}, vol.~20, no.~11, pp.~1254--1259, 1998.

\bibitem{123}
T.-T. Ng, S.-F. Chang, J.~Hsu, and M.~Pepeljugoski, ``Columbia photographic
  images and photorealistic computer graphics dataset,'' {\em Columbia
  University, ADVENT Technical Report}, pp.~205--2004, 2005.

\bibitem{ref53}
D.~Lowe, ``Object recognitions using local scale-invariant features,'' in {\em
  IEEE International Conference on Computer Vision}, 1999.

\bibitem{ref54}
M.~Douze, H.~Jegou, H.~Sandhawalia, L.~Amsaleg, and C.~Schmid, ``Evaluation of
  gist descriptors for web-scale image search,'' in {\em ACM International
  Conference on Image and Video Retrieval}, 2009.

\bibitem{ref55}
N.~Dalal and B.~Triggs, ``Histograms of oriented gradients for human
  detection,'' in {\em IEEE Computer Vision and Pattern Recognition}, 2005.

\bibitem{124}
A.~Khosla, A.~S. Raju, A.~Torralba, and A.~Oliva, ``Understanding and
  predicting image memorability at a large scale,'' in {\em IEEE International
  Conference on Computer Vision}, pp.~2390--2398, 2015.

\bibitem{ref56}
N.~Ahmed, T.~Natarajan, and K.~R. Rao, ``Discrete cosine transform,'' {\em IEEE
  Transactions on Computers}, vol.~23, pp.~90--93, 1974.

\bibitem{uzairDCT}
M.~Uzair, A.~Mahmood, and A.~S. Mian, ``Hyperspectral face recognition using
  3d-dct and partial least squares.,'' in {\em British Machine Vision
  Conference (BMVC)}, vol.~1, p.~10, 2013.

\bibitem{125}
D.~D. Lewis, ``Naive (bayes) at forty: The independence assumption in
  information retrieval,'' in {\em European conference on machine learning},
  pp.~4--15, Springer, 1998.

\bibitem{126}
L.~Zhang and H.~Li, ``Sr-sim: A fast and high performance iqa index based on
  spectral residual,'' in {\em IEEE international conference on image
  processing}, pp.~1473--1476, 2012.

\bibitem{uzair2016blind}
M.~Uzair and A.~Mian, ``Blind domain adaptation with augmented extreme learning
  machine features,'' {\em IEEE Transactions on Cybernetics}, vol.~47, no.~3,
  pp.~651--660, 2017.

\bibitem{uzair2018representation}
M.~Uzair, F.~Shafait, B.~Ghanem, and A.~Mian, ``Representation learning with
  deep extreme learning machines for efficient image set classification,'' {\em
  Neural Computing and Applications}, vol.~30, no.~4, pp.~1211--1223, 2018.

\bibitem{ref57}
Y.~Lifang, Q.~Sijun, and Z.~Huan, ``Feature selection algorithm for
  hierarchical text classification using kullback-leibler divergence,'' in {\em
  IEEE Int. Conf. on Cloud Computing and Big Data Analysis}, 2017.

\bibitem{ref62}
R.~Schadewald, ``Moon and spencer and the small universe,'' {\em Evolution
  Journal}, vol.~2, no.~2, pp.~20--22, 1981.

\bibitem{ref63}
R.Iordache, A.~Beghdadi, and P.~V. de~Lesegno, ``Pyramidal perceptual filtering
  using moon and spencer contrasr,'' in {\em Proceedings of International
  Conference on Image Processing}, 2001.

\bibitem{ref61}
K.~Liu and J.~Yang, ``Recognition of people reoccurences using bag-of-features
  representation and support vector machine,'' in {\em Chinese Conference on
  Pattern Recognition}, 2009.

\bibitem{ref64}
J.~Sanchez, F.~Perronnin, T.~Mensink, and J.~Verbeek, ``Image classification
  with the fisher vector: Theory and practice,'' {\em International Journal of
  Computer Vision}, 2013.

\bibitem{ref65}
A.~Faheema and S.~Rakshit, ``Feature selection using bag-of-visual-words
  representation,'' in {\em IEEE 2nd International Conference on Advanced
  Computing Conference (IACC)}, 2010.

\bibitem{mclachlan2007algorithm}
G.~J. McLachlan and T.~Krishnan, {\em The EM algorithm and extensions},
  vol.~382.
\newblock John Wiley \& Sons, 2007.

\bibitem{ref68}
J.~Lewis, ``Creation by refinement: a creativity paradigm for gradient descent
  learning networks,'' in {\em Int. Conf. on Neural Networks}, 1988.

\bibitem{ref69}
Takefuji, ``Parallel distributed gradient descent and ascent methods,'' in {\em
  International Joint Conference on Neural Networks (IJCNN)}, 1989.

\bibitem{128}
R.~Sun, Z.~Lian, Y.~Tang, and J.~Xiao, ``Aesthetic visual quality evaluation of
  chinese handwritings,'' in {\em International Joint Conference on Artificial
  Intelligence}, 2015.

\bibitem{datta2006studying}
R.~Datta, D.~Joshi, J.~Li, and J.~Z. Wang, ``Studying aesthetics in
  photographic images using a computational approach,'' in {\em European
  conference on computer vision}, pp.~288--301, Springer, 2006.

\bibitem{129}
J.~Xiang and G.~Zhu, ``Joint face detection and facial expression recognition
  with mtcnn,'' in {\em International Conference on Information Science and
  Control Engineering}, pp.~424--427, 2017.

\bibitem{ref17}
Y.~Kao, K.~Huang, and S.~Maybank, ``Hierarchical aesthetic quality assessment
  using deep convolutional neural networks,'' {\em Image Communication
  Journal}, vol.~47, pp.~500--510, 2016.

\bibitem{ref27a}
L.~Li, H.~Zhu, S.~Zhao, G.~Ding, H.~Jiang, and A.~Tan, ``Personality driven
  multi-task learning for image aesthetic assessment,'' in {\em International
  Conference on Multimedia and Expo}, pp.~430--435, 2019.

\bibitem{ref18}
S.~Bianco, L.~Celona, P.~Napoletano, and R.~Schettini, ``Predicting image
  aesthetics with deep learning,'' in {\em International Conference Advanced
  Concepts for Intelligent Vision Systems}, pp.~117--125, 2016.

\bibitem{ref29a}
L.~Li, H.~Zhu, S.~Zhao, G.~Ding, and W.~Lin, ``Personality-assisted multi-task
  learning for generic and personalized image aesthetics assessment,'' {\em
  IEEE Transactions on Image Processing}, vol.~29, pp.~3898--3910, 2020.

\bibitem{ref19}
S.~Kong, X.~Shen, Z.~Lin, R.~Mech, and C.~Fowlkes, ``Photo aestheticsranking
  network with attributes and content adaptation,'' in {\em Eurpeon Conference
  on Computer Vision}, 2016.

\bibitem{ref20}
X.~Lu, Z.~Lin, H.~Jin, J.~Yang, and J.~Z. Wang, ``Rating image aesthetics using
  deep learning,'' {\em IEEE Tranactions on Multimedia}, vol.~17, no.~11,
  pp.~2011--2034, 2014.

\bibitem{ref21}
L.~Mai, H.~Jin, and F.~Liu, ``Composition-preserving photo aesthetics
  assessment,'' in {\em IEEE Conference on Computer Vision and Pattern
  Recognition (CVPR)}, 2016.

\bibitem{ref22}
Y.~Wang, Y.~Li, and F.~Porikli, ``Fine-tuning convolutional neural networks for
  visual aesthetics,'' in {\em 2016 23rd International Conference on Pattern
  Recognition (ICPR)}, 2016.

\bibitem{ref24}
Z.~Wang, F.~Dolcos, D.~Beck, S.~Chang, and T.~S. Huang, ``Brain-inspired deep
  networks for image aesthetics assessment,'' {\em CoRR}, vol.~abs/1601.04155,
  2016.

\bibitem{ref25a}
Z.~Liu, Z.~Wang, Y.~Yao, L.~Zhang, and L.~Shao, ``Deep active learning with
  contaminated tags for image aesthetics assessment,'' {\em IEEE Transactions
  on Image Processing}, 2018.

\bibitem{ref30a}
J.~McCormack and A.~Lomas, ``Deep learning of individual aesthetics,'' {\em
  Neural Computing and Applications}, vol.~33, no.~1, pp.~3--17, 2021.

\bibitem{ref31a}
X.~Zhang, X.~Gao, L.~He, and W.~Lu, ``Mscan: Multimodal self-and-collaborative
  attention network for image aesthetic prediction tasks,'' {\em
  Neurocomputing}, vol.~430, pp.~14--23, 2021.

\bibitem{khan2019dimension}
S.~D. Khan, H.~Ullah, M.~Ullah, F.~A. Cheikh, and A.~Beghdadi, ``Dimension
  invariant model for human head detection,'' in {\em 2019 8th European
  Workshop on Visual Information Processing (EUVIP)}, pp.~99--104, IEEE, 2019.

\bibitem{khan2019person}
S.~D. Khan, H.~Ullah, M.~Ullah, N.~Conci, F.~A. Cheikh, and A.~Beghdadi,
  ``Person head detection based deep model for people counting in sports
  videos,'' in {\em IEEE International Conference on Advanced Video and Signal
  Based Surveillance}, pp.~1--8, 2019.

\bibitem{ref15}
Y.~Zhou, G.~Li, and Y.~Tan, ``Computational aesthetics of photo quality
  assessment and classification based on artificial neural network with deep
  learning methods,'' {\em Int. Journal of Signal Processing, Image Processing
  and Pattern Recognition}, vol.~8, no.~1, pp.~173--282, 2015.

\bibitem{ref26a}
X.~Fu, J.~Yan, and C.~Fan, ``Image aesthetics assessment using composite
  features from off-the-shelf deep models,'' in {\em IEEE International
  Conference on Image Processing}, pp.~3528--3532, 2018.

\bibitem{ref16}
Y.~Kao, R.~He, and K.~Huang, ``Deep aesthetic quality assessment with semantic
  information,'' {\em IEEE Transactionson on Image Processing}, vol.~26, no.~3,
  pp.~1482--1495, 2017.

\bibitem{ref79}
A.~Giusti, D.~C. Ciresan, J.~Mascai, L.~M. Gambardella, and J.~Schmidhuber,
  ``Fast image scanning with deep max-pooling convolutional neural networks,''
  in {\em IEEE Int. Conference on Image Processing}, 2013.

\bibitem{ref80}
M.~Yang, B.~Li, H.~Fan, and Y.~Jiang, ``Randomized spatial pooling in deep
  convolutional neural networks for scene recognition,'' in {\em IEEE
  International Conference on Image Processing (ICIP)}, 2015.

\bibitem{ref81}
L.~Liu, C.~Shen, and A.~van~den Hengel, ``The treasure beneath convolutional
  layers: Cross-convolutional-layer pooling for image classification,'' in {\em
  IEEE Conference on Computer Vision and Pattern Recognition (CVPR)}, 2015.

\bibitem{ref82}
Y.~Jia, E.~Shelhamer, J.~Donahue, S.~Karayev, J.~Long, R.~Girshick,
  S.~Guadarrama, and T.~Darell, ``Caffe: Convolutional architecture for fast
  feature embedding,'' in {\em IEEE Conference on Computer Vision and Pattern
  Recognition (CVPR)}, 2014.

\bibitem{ref83}
J.~Wang, Y.~Li, Z.~Miao, Y.~Yu, and G.~Tao, ``Euclidean output layer for
  discriminative feature extraction,'' in {\em 2nd International Conference on
  Image, Vision and Computing (ICIVC)}, 2017.

\bibitem{ref84}
S.~Tao, T.~Zhang, J.~Yang, X.~Wang, and W.~Lu, ``Bearing fault diagnosis mathod
  based on stacked autoencoder and softmax regression,'' in {\em 34th Chinese
  Control Conference (CCC)}, 2015.

\bibitem{130}
S.~Kong, X.~Shen, Z.~Lin, R.~Mech, and C.~Fowlkes, ``Photo aesthetics ranking
  network with attributes and content adaptation,'' in {\em European Conference
  on Computer Vision}, pp.~662--679, Springer, 2016.

\bibitem{ref88}
A.~Krizhevsky, T.~Sutskever, and G.~E. Hinton, ``Imagenet classification with
  deep convolutional neural networks,'' in {\em Proceedings of Int. Conf. on
  Neural Information Processing Systems}, pp.~1097--1105, 2012.

\bibitem{ref89}
I.~Melekhov, J.~Kannala, and E.~Rahtu, ``Siamese network features for image
  matching,'' in {\em 23rd International Conference on Pattern Recognition
  (ICPR)}, 2016.

\bibitem{ref90}
Y.~Qi, Y.-Z. Song, H.~Zhang, and J.~Liu, ``Sketch-based image retrieval via
  siamese convolutional neural network,'' in {\em IEEE International Conference
  on Image Processing (ICIP)}, 2016.

\bibitem{131}
C.~Gao, Y.~Du, J.~Liu, L.~Yang, and D.~Meng, ``A new dataset and evaluation for
  infrared action recognition,'' in {\em CCF Chinese Conference on Computer
  Vision}, pp.~302--312, Springer, 2015.

\bibitem{ullah2019two}
H.~Ullah, S.~D. Khan, M.~Ullah, F.~A. Cheikh, and M.~Uzair, ``Two stream model
  for crowd video classification,'' in {\em European Workshop on Visual
  Information Processing}, pp.~93--98, 2019.

\bibitem{132}
L.~Wang, S.~Guo, W.~Huang, and Y.~Qiao, ``Places205-vggnet models for scene
  recognition,'' {\em arXiv preprint arXiv:1508.01667}, 2015.

\bibitem{133}
N.~L{\'e}onard, S.~Waghmare, Y.~Wang, and J.-H. Kim, ``rnn: Recurrent library
  for torch,'' {\em arXiv preprint arXiv:1511.07889}, 2015.

\bibitem{ref23}
K.-H. Lu, K.-Y. Chang, and C.-S. Chen, ``Image aesthetic assessment via deep
  semantic aggregation,'' in {\em IEEE Global Conference on Signal and
  Information Processing}, 2016.

\bibitem{135}
B.~Zhou, A.~Lapedriza, A.~Khosla, A.~Oliva, and A.~Torralba, ``Places: A 10
  million image database for scene recognition,'' {\em IEEE transactions on
  pattern analysis and machine intelligence}, vol.~40, no.~6, pp.~1452--1464,
  2017.

\bibitem{ref28a}
Q.~Chen, W.~Zhang, N.~Zhou, P.~Lei, Y.~Xu, Y.~Zheng, and J.~Fan, ``Adaptive
  fractional dilated convolution network for image aesthetics assessment,'' in
  {\em Proceedings of the IEEE/CVF Conference on Computer Vision and Pattern
  Recognition}, pp.~14114--14123, 2020.

\bibitem{Assessing2011Luca}
L.~Marchesotti, F.~Perronnin, D.~Larlus, and G.~Csurka, ``Assessing the
  aesthetic quality of photographs using generic image descriptors,'' in {\em
  2011 international conference on computer vision}, pp.~1784--1791, IEEE,
  2011.

\bibitem{ref5}
R.~Gadde and K.~Karlapalem, ``Aesthetic guideline driven photography by
  robots,'' in {\em Proceedings of the International Joint Conference on
  Artificial Intelligence}, 2011.

\end{thebibliography}
\end{document}